# A comprehensive cross-language framework for harmful content detection with the aid of sentiment analysis

Mohammad Dehghani
School of Electrical and Computer Engineering,
University of Tehran,
Tehran, Iran
dehghani.mohammad@ut.ac.ir

**Abstract**
In today's digital world, social media plays a significant role in facilitating communication and content sharing. However, the exponential rise in user-generated content has led to challenges in maintaining a respectful online environment. In some cases, users have taken advantage of anonymity in order to use harmful language, which can negatively affect the user experience and pose serious social problems. Recognizing the limitations of manual moderation, automatic detection systems have been developed to tackle this problem. Nevertheless, several obstacles persist, including the absence of a universal definition for harmful language, inadequate datasets across languages, the need for detailed annotation guideline, and most importantly, a comprehensive framework. This study aims to address these challenges by introducing, for the first time, a detailed framework adaptable to any language. This framework encompasses various aspects of harmful language detection. A key component of the framework is the development of a general and detailed annotation guideline. Additionally, the integration of sentiment analysis represents a novel approach to enhancing harmful language detection. Also, a definition of harmful language based on the review of different related concepts is presented. To demonstrate the effectiveness of the proposed framework, its implementation in a challenging low-resource language is conducted. We collected a Persian dataset and applied the annotation guideline for harmful detection and sentiment analysis. Next, we present baseline experiments utilizing machine and deep learning methods to set benchmarks. Results prove the framework's high performance, achieving an accuracy of 99.4% in offensive language detection and 66.2% in sentiment analysis.

## 1. Introduction

Social media provide a means of communicating with people worldwide [1]. With the advent of smartphones, people are increasingly using social media [2] and users have access to a discussion space [3]. The use of social media has become an integral part of every individual's daily life, enabling them to communicate with one another in a virtual environment, thereby allowing them to express their views and opinions freely [4]. The lack of disincentives within society has created a situation where people express their ideas on different matters, regardless of their level of media literacy or the accuracy of their viewpoints [5]. As a result, social media has significantly affected global language, culture, interactions, and communication [6, 7].
Users tend to misuse the anonymity provided by online social media, taking advantage of it, and engaging in behavior that is not socially acceptable in the real world [8]. The anonymity of the internet is allowing individuals to display harmful language, which is on the rise. The absence of proper legal proceedings to tackle harmful language on social media has become a serious problem [9]. These types of behavior has a negative effect on user's mental and physical health [10, 11]. Also, lead to irreparable consequences (such as anxiety, depression, self-harm, suicide, etc.) [12]. Children and adolescents who are exposed to harmful language at a young age may suffer negative consequences. Experiencing negative emotions, such as anger, fear, and sadness, is more likely for children exposed to online harmful language [13]. There is a possibility that it may affect their attitudes, behaviors, and social interactions, contributing to a cycle of negative behavior [14, 15].

With the increasing crime due to harmful messages received by users, it is very important to have some form of a filter that could prevent the attacking user from sending such harmful content which could destroy the peace of individuals and society [16]. It is essential to tackle this problem by detecting and eliminating harmful content as soon as possible and making the online platform more secure and safe [17]. The principal purpose of identifying harmful language is early detection, protection, prevention, discouraging, and amendment [18, 19].

Social media platforms are under considerable pressure to identify and remove harmful language in a timely manner [20]. Companies in the social media, online gaming, and dating industries have been focused on harmful language detection and elimination as a means of protecting their users. In most cases, such companies have a Trust and Safety department that defines and enforces their content policies, as well as develops tools for identifying instances of harmfulness on their websites. A critical component of creating safe and equitable online spaces is content moderation, which involves monitoring and reviewing user-generated content in order to ensure compliance with the user agreement, community guidelines, and legal requirements [21].

Therefore, a strategy to deal with harmful language involves manually monitoring and moderating user-generated content [22]. Moderators are employed on these platforms to maintain a respectful tone and to enforce the platform's discussion rules, including the prohibition of harmful language [3]. It is, however, difficult to apply manual methods of moderation and intervention to the web due to its rapid rate of new data generation [22]. Therefore, it has become increasingly important to use (semi-) automatic methods in order to identify such behavior in recent years. A tool that detects and prevents harmful communication can be beneficial to governments and social media platforms since it can lead to severe harm to society [23].

Increasing threats to the online platforms (such as fake news, toxic discourse, violence, and harmful language) have led to an increased demand in the natural language processing (NLP) community for tasks addressing social and ethical issues [24]. The harmful language detection can be accomplished by using large volumes of online data and applying appropriate NLP and machine/deep learning techniques. Detecting harmful language is a classification task in which the model determines whether a text contains harmful language or not.

However, it is challenging to perform this task due to the fact that there are so many ways to attack a person or group, both explicitly and implicitly [25]. Therefore, sentiment analysis can be useful in detecting harmful language. Sentiment analysis is another classification task in NLP that has gained considerable attention in recent years [26]. Sentiment analysis involves analyzing a text to determine whether it expresses a positive, negative, or neutral sentiment [27]. By identifying negative sentiments in text, sentiment analysis algorithms can help flag potentially harmful content, as harmful language tends to exhibit negative sentiment characteristics. This alignment between harmful language and negative sentiment analysis underscores the relevance and effectiveness of sentiment analysis techniques in harmful language detection [28].

It should be noted that most of NLP systems are designed to analyze languages in formal settings with correct grammar. As a result, analyzing comments that are provided by users presents a challenge [29]. Also, in spite of existing models, it may not be possible to make people entirely immune to harmful language due to the difficulty in determining what is "harmful". Various research gaps exist regarding sarcasm, complex words and sentences that contain grammatical errors, as well as punctuation [16]. Hence, there is a need for further research and provide more accurate solutions. As a major innovation of this study, we proposed a comprehensive framework for harmful language detection which consists of necessary elements to develop such systems. Within this framework, we provided a detailed annotation guideline to ensure the data is labeled correctly. In addition, we emphasize the importance of identifying the source of harmfulness in order to comprehend why a text is harmful.

As a result of this paper, the following contributions have been made:

(1) Developing a cross-language framework for harmful language detection.
(2) Providing an accurate definition for harmful language by reviewing different definitions (offensive, abuse, hateful, etc.).
(3) Designing a complete guideline for annotation of corpus in the context of harmful language and sentiment analysis.
(4) Identification of the source of offensiveness in order to provide insight into the logic behind the results.
(5) To demonstrate the efficacy of our proposed framework, a dataset for a low resource language is collected, data is labeled using our annotation guidelines, baseline models are implemented and evaluate the results are evaluated.
(6) Providing a list of harmful words for targeted low resource language.

The rest of paper organized as following. In Section 2, previouse works are reviewed. Section 3 provide a survey about datasets. In Section 4, existing challenges are outlined. Section 5 present a diffinition for harmful detection and its types. Section 6, we introduce proposed framework and its components. Section 7 presents an implementation to assess the framework's efficacy. Section 8 comprises discussion and comparison with several previous studies. Finally, Section 9 concludes the paper.

**Disclaimer:** The document may contain harm language, which do not reflect the authors' views. The goal is to combat such forms of language.

## 2. Related works

The literature provides a variety of approaches to identify offensive texts using machine learning and deep learning as showed in Table 1. Yuan et al. [30] noted that other hate speech studies rely on custom features, user data, or meta-data specific to the platform, which limit their ability to generalize to new data sets. As a result, they attempted to develop a general-purpose embedding and detection system for hate. To address the dataset problem, they use smaller, unrelated data sets to learn jointly (Waseem data set [31] and Davidson data set [32]). A model based on transfer learning methods is proposed. A fully connected layer and a softmax activation layer are employed to classify hate. Finally, the author introduces the Map of Hate, which illustrates hateful content. The Map of Hate can be used to distinguish between different types of hate and explain what makes a text hateful.
The aim of Molero et al. [33] was to detect offensive language in Spanish texts by taking into account the unbalanced nature of the data. In order to perform the analysis, stochastic gradient descent (SGD), support vector machine (SVM), random forest, gradient boosting, and adaboost were used with Bag-of-Words as a data representation**.** Additionally, deep learning models, such as convolutional neural network (CNN), Bi-LSTMs, and transformers, were considered. According to the evaluation of the OffendES dataset [34], transformer-based models achieved the best results and improved previous results by 6,2%.
A method based on BERT was proposed by Chavan et al. [35] for detecting offensive language in the language Marathi, which is a low resource language. The performance of MuRIL, MahaTweetBERT, MahaTweetBERT-Hateful, and MahaBERT was compared. HASOC 2022 test set is considered for training and evaluation aloge with data augmented from other Marathi hate speech corpus HASOC 2021 and L3Cube-MahaHate. The best result was achieved by MahaTweetBERT with an F1 score of 98.43.

To address the multilingual offensive language detection problem, El-Alami et al. [36] developed a transformer-based model. In order to deal with multilingualism, two different approaches were used. The first approach used was joint-multilingual, which involves developing one classification system for several languages. Another technique is translation-based, which involves transforming all texts into a single universal language before they are classified. Using the SOLID and an Arabic dataset, the translation-based method with Arabic BERT achieves an F1-score of 93% and accuracy of 91%.

Rizwan et al. [37] provide a lexicon of hateful words in Urdu and create an annotated dataset called RUHSOLD consisting of 10012 tweets. The proposed BERT+CNN-gram for hate-speech and offensive detection has the highest results with accuracy and F1-score of 90%.

Akhter et al. [38] performed abusive language detection in Urdu and Roman Urdu comments. In order to classify the data, they used five machine learning models and four deep learning models. The natural language constructs, the English-like nature of Roman Urdu script, and the Nastaleeq style of Urdu present challenges in processing and categorizing the comments. Convolutional neural networks outperform the other models and achieve accuracy of 96.2% and 91.4% for Urdu and Roman Urdu, respectively.
The focus of Risch et al. [3] was the explanation of classification results. A multinomial naive Bayes classifier was employed, which offers explanations in the form of conditional probabilities. Other classifiers include TF-IDF and Glove vector representations, which are integrated with SVM and LSTM models, and are complemented by two explanation methods, LRP [39] and LIME [40]. Models are evaluated using the toxic comments dataset, which contains approximately 220,000 comments. There are six classes in this dataset, including toxic, severe toxic, insult, threat, obscene, and identity hate. The results indicate that LSTM with LRP has the highest Explanatory Power Index (EPI), equal to 99.67%.

Table 1: Summary of previous studies.

| **Article** | **Year** | **Dataset** | **Language** | **Number of samples** | **Text representations** | **Classifiers** | **Advantage** | **Category** |
|---|---|---|---|---|---|---|---|---|
| [30] | 2023 | Waseem dataset, Davidson dataset | English | 15,216 tweets from Waseem, 22,304 tweets from Davidson | ELMo 5.5B, Glove | BiLSTM | Using two datasets jointly, identify types of hate, explain the classification results | Hate speech |
| [33] | 2023 | OffendES | Spanish | 30,416 comments from X, Instagram, and YouTube | BOW, TF-IDF | SGD, SVM, RandomForest, GradientBoosting, AdaBoost, CNNs, Bi-LSTMs, Transformers | Addressing the unbalanced nature of data | Offensive Language |
| [35] | 2022 | HASOC 2022, HASOC 2021 + HASOC 2022, HASOC 2021 + HASOC 2022 + MahaHate | Marathi | HASOC 2022 = 3096, HASOC 2021 = 1874, MahaHate = 6250 tweets | BERT | BERT | Low resource language, data augmentation | Offensive Language |
| [36] | 2022 | SOLID | English and Arabic | 6000 English tweets from the SOLID, 7800 Arabic tweets | BERT | BERT | Multilingual | Offensive Language |
| [38] | 2021 | GitHub dataset, Urdu Offensive Dataset | Roman Urdu, Urdu | 10000 for Roman Urdu, 2171 for Urdu | TF-IDF, BoW, word embedding | NB, SVM, IBK, Logistic, JRip, CNN, LSTM, Bi-LSTM, CLSTM | Low resource language | Abusive language |
| [3] | 2020 | Toxic comments dataset | English | 220,000 comments | TF-IDF, Glove | Multinomial Naive Bayes, SVM, LSTM | Using explanatory models | Offensive Language |
| [37] | 2020 | RUHSOLD | Urdu | 10012 tweets | BERT | CNN-gram | Create lexicon for low resource language | Hate-speech and offensive detection |

## 3. Datasets

Researchers collected datasets from various online platforms and annotated them for a variety of languages. We categorized datasets into the following groups according to their resource languages:

1) **High resource languages:** Languages with benchmark datasets annotated grammatical, and semantically precision in large scale commonly are cross-domain, and topic independent [41-43].
2) **Low resource languages:** Languages with benchmark datasets annotation are developing. Their size is medium to large. Commonly context, domain, and topic dependent [44, 45].
3) **Multilingual resources:** There are studies that take into account both high level and low level languages and provide datasets corresponding to various languages.

Several study from each category are reviewed in the following and more datasets are introduced in the Tables 2,3, and 4.

### 3.1. High resource languages

Those languages that have a considerable amount of linguistic data, research, and technology resources are considered high-resource languages. They have a wide range of annotated corpora [46]. Languages in this category include English, Chinese, Japanese, Russian, as well as a number of western European languages [47] such as French, Italian, and Spanish [48], which have rich resources in harmful language domain.
Approximately 8,500 tweets were collected by Wiegand et al. [49] for the purpose of detecting offensive language in Germany. A heuristic selection of 100 users who regularly post offensive tweets was carried out in order to generate a more diverse vocabulary set. The data was manually annotated by three native German speakers who organized the GermEval 2018 shared task.
According to Zampieri et al., [50] most existing datasets consider only one type of offensive language such as cyberbullying or hate speech. Therefore, they establish a dataset called OLID based on a three-level hierarchical annotation schema that was performed manually. In the first level, offensive language is detected and the tweets are classified as offensive or not offensive. The second level of the process involves determining the type of offense (targeted or untargeted insult). As a final step, the target of insults is determined, which can be either an individual, a group, or other.
A limitation of OLID, according to Rosenthal et al. [51], is its limited size, particularly for some of the low-level categories, posing a challenge when training robust models. The authors used OLID as a seed dataset and presented SOLID as a dataset for offensive language identification. There are over 12 million tweets in Solid that contain both explicit (such as curse words) as well as implicit (such as racial slurs) offensive content.
Davidson et al. [32] collected tweets containing hate speech keywords using a hate speech lexicon. Hatebase.org compiles a list of words that internet users have identified as hate speech. Based on crowdsourcing, data were categorized into hate speech, offensive language, and neither. Users labelled the items based on both the lexicon and context of their use, and the final labels were determined by majority vote.
An overview of high resource language datasets is provided in Table 2.

Table 2: High resource language datasets.

| **Article** | **Year** | **Dataset Name** | **Language** | **Number of samples** | **Source of text** | **Category** |
|---|---|---|---|---|---|---|
| [52] | 2022 | DeTox | Germany | 10,278 | X | Offensive Language |
| [53] | 2022 | COLD | Chinese | 37,480 | Zhihu, Weibo | Offensive Language |
| [54] | 2022 | ChileOLD | Spanish | 9834 | X | Hate speech |
| [55] | 2022 | SWSR | Chines | 8,969 | Weibo | Sexism |
| [56] | 2021 | - | Russian | 15,797 | X | Abusive language |
| [57] | 2021 | - | Japanese | 9,449,645 (words) | X | Hate speech |
| [51] | 2020 | SOLID | English | 12,000,000 | X | Offensive Language |
| [58] | 2020 | SOCC | English | 663,173 | News Site Comments (The Globe and Mail and The New York Times) | Toxic |
| [59] | 2020 | AbuseEval v1.0 | English | 14,100 | X | Abusive language |
| [60] | 2020 | COLA | Chines | 18,707 | COLA-Youtube, Weibo | Offensive Language |
| [61] | 2020 | - | English | 9,093,037 | X | Offensive Language |
| [50] | 2019 | OLID | English | 14,101 | X | Offensive Language |
| [62] | 2019 | - | Italian | 4,000 | X | Hate speech |
| [63] | 2019 | - | French | 4,014 | X | Hate speech |
| [22] | 2018 | - | English | 15,000 | Facebook | Aggression |
| [49] | 2018 | - | Germany | 8,541 | X | Offensive Language |
| [64] | 2018 | - | English | 80,000 | News Sites | Offensive Language, Hate Speech, Cyberbullying |
| [32] | 2017 | - | English | 24,802 | X | Hate speech |
| [65] | 2017 | - | English | 100,000 | X | Hate speech |
| [66] | 2017 | - | English | 24,840 | Youtube | Violence |
| [67] | 2017 | - | English | 74,874 | X | Sexism |
| [68] | 2017 | | English | 35,000 | X | Harassment |
| [69] | 2017 | - | English | 1007 | News Site comments (MediaGist) | Flame |
| [69] | 2017 | - | Italian | 649 | News Site comments (MediaGist) | Flame |
| [69] | 2017 | - | Germany | 1122 | News Site comments (MediaGist) | Flame |
| [69] | 2017 | - | French | 487 | News Site comments (MediaGist) | Flame |
| [70] | 2016 | - | English | 3,325,636 | News Site comments (Yahoo! Finance and news) | Abusive language |
| [71] | 2016 | - | English | 16,914 | X | Hate speech |
| [72] | 2015 | - | English | 3,165,000 | Instagram | Cyberbullying |

### 3.2. Low resource languages

There is a lack of linguistic data, research, and technological resources in low-resource languages. They are less computerized, less privileged, and rarely taught [47, 73]. Among these languages are Persian (also known as Farsi), Korean [74], and Bengali, Spanish, Hindi, Turkish and Greek [75].

A number of harmful language datasets collected for the Persian language. Dehghani et al. [76] obtained a dataset from X for abusive language detection. A total of 33,338 tweets were collected, of which 10% were abusive and 90% were not abusive. A variety of machine learning and deep learning methods were used for classification. BERT outperformed other models with an accuracy of 97.7% and an F1-score of 99.3%. This study also reported train and test times, which are important criteria as detecting harmful language as soon as possible is paramount. This will result in fewer users being exposed to such inappropriate content. In comparison with other models, logistic regression has the shortest test time of 0.054 seconds with a 95.8% accuracy.

Ataei et al. [77] presented the Pars-OFF dataset for offensive detection in Persian including 10,563 tweets. A three-layered annotated corpus was developed following OLID guidelines and customized it for Persian. The first level consists of two classes, offensive and non-offensive. The second level of analysis is to determine whether the offensive tweets have a target or not. At the third level, targeted tweets are classified into three categories: individual, group, and other. Dataturks [78] and Kili-technology [79] crowdsourcing platforms were used by three Persian native speakers to annotate the data. To resolve the imbalance in data collection, a dataset expansion was performed by using similarity-based and keyword-based approaches. A Keyword-based classifier, a combination of Naive Bayes and SVM, BERT, Bidirectional LSTM, and BERT + fastText were used for offensive language detection. The F1-Macro score for the BERT+fastText model was 89.57.

Using the TWINT crawler and a list of keywords, Kebriaei et al. [80] collected 38,000 tweets. They were tasked with detecting offensive and hate speech. For classification, SVM, Logistic Regression, mBERT, ParsBERT, XLM-RoBERTa, and BERTweet-FA were used. BERTweet-FA achieved best results with precision of 90.3% on the original dataset and 90.2% on the revised dataset.

Table 3 present a summary of low resource language datasets.

Table 3: Low resource language datasets.

| **Article** | **Year** | **Dataset Name** | **Language** | **Number of samples** | **Source of text** | **Category** |
|---|---|---|---|---|---|---|
| [80] | 2023 | - | Persian | 38,000 | X | Offensive Language, Hate Speech |
| [81] | 2023 | OMCD | Moroccan | 80,24 | YouTube | Offensive Language |
| [82] | 2023 | PerBOLD | Persian | 28164 | Instagram | Offensive Language |
| [77] | 2022 | Pars-OFF | Persian | 10,563 | X | Offensive Language |
| [83] | 2022 | KOLD | Korean | 40,429 | NAVER news, YouTube | Offensive Language |
| [76] | 2021 | - | Persian | 33,338 | X | Abusive language |
| [84] | 2021 | POLID (for Persian OLID) | Persian | 4,988 | X, Instagram, users' reviews on different Iranian web applications | Offensive Language |
| [85] | 2021 | - | Persian | 6,000 | X | Offensive Language |
| [86] | 2021 | - | Urdu | 3,500 | X | Abusive language |
| [61] | 2020 | - | Arabic | 10,000 | X | Offensive Language |
| [61] | 2020 | - | Danish | 3,290 | X | Offensive Language |
| [61] | 2020 | OGTD | Greek | 10,287 | X | Offensive Language |
| [61] | 2020 | - | Turkish | 35,000 | X | Offensive Language |
| [87] | 2020 | TurkishOLD | Turkish | 34,792 | X | Offensive Language |
| [88] | 2019 | PolEval | Polish | 10,041 | X | Cyberbullying, Hate Speech |
| [89] | 2019 | HinHD | Hindi | 8,192 | X, Facebook, WhatsApp | Hate Speech |
| [90] | 2018 | - | Mexican | 11,000 | X | Aggression |
| [91] | 2018 | - | Amharic | 6,120 | Facebook | Hate Speech |
| [92] | 2017 | - | Greek | 1,600,000 | News sites comments | Offensive Language |
| [69] | 2017 | - | Czech | 1812 | News Site comments (MediaGist) | Flame |

### 3.3. Multilingual resources

The objective of multilingual resources is to create models that can detect harmful language regardless of its language of origin [26]. It is also intended to make use of existing English datasets to improve the performance of other language systems [75].

A multilingual dataset was generated by Chakravarthi et al. [26] using YouTube comments in Tamil-English, Kannada-English, and Malayalam-English. Manual annotations were performed by volunteer annotators for the purpose of sentiment analysis and the identification of offensive languages. Data were annotated by students from several universities and family members of the authors using Google Form. At least three individuals annotated each form. Annotations were made by two persons for each sentence, and as long as both agreed, the data was included in the collection. The sentence was otherwise annotated by a third person. In the event that the three annotators disagreed, two additional annotators were consulted.

A dataset from Facebook was provided by Kumar et al. [22] to identify aggression, and the texts were classified into overtly aggressive, covertly aggressive, and non-aggressive categories. For testing, 30 teams participated in the Aggression Identification Shared Task at TRAC-1 using data from Facebook and X.

There is a list of datasets for multilingual resources in Table 4.

Table 4: Multilingual resource datasets.

| Article | Year | Dataset Name | Language | Number of samples | Source of text | Category |
|---|---|---|---|---|---|---|
| [26] | 2022 | DravidianCodeMix | Tamil-English, Kannada-English, Malayalam-English | More than 60,000 | YouTube comments | Offensive Language |
| [93] | 2020 | XHATE-999 | English, Albanian, Croatian, German, Russian, Turkish | 999 | GAO, TRAC, and WUL datasets | Abusive language |
| [94] | 2019 | HatEval | English, Spanish | 19,600 | X | Hate speech |
| [22] | 2018 | - | English-Hindi | 19,337 | Facebook, X | Aggression |
| [95] | 2018 | HEOT | Hindi-English | 3,679 | X | Offensive Language |

Various data sets have been provided for the detection of harmful language in different languages. Many of these datasets were annotated in a similar manner to the articles mentioned above and most of them followed OLID guidelines. The majority of datasets are created in English, with X being the primary source for data collection. For high resource languages such as English, data is collected from various social media platforms. Low resource languages receive less attention and have fewer datasets available. In recent studies, more researchers aimed to provide datasets for multilingual and low resource languages.

## 4. Harmful language detection challenges

It is crucial to design an automated system for harmful language detection on online platforms, since this can have a variety of negative effects on users. There are a number of datasets and methods available in this regard, but the subject still faces some serious challenges. Several challenges have been identified in the following, and this study aims to address them and provide appropriate solutions.

**(1) What is harmful language?**

The problem we encountered as we reviewed more and more studies is that there is no universal definition of harmful language. It appears that there is no clear distinction between different types of harmful language. The importance of this is that we must be aware of what should be counted as harmful and what should be considered non-harmful during the annotation process. In other words, this represents the basis for defining the rules of annotation. Consequently, it is crucial to define harmful language properly and differentiate it from other concepts such as offensive, abuse, hate speech, etc. We address this challenge in Section 5.

**(2) The lack of a comprehensive framework for harmful language detection**
In nearly all fields of science, frameworks are critical for organizing empirical research and developing theories [96]. In order to create harmful language detection systems, a framework can provide a guideline by determining what requirements must be met. This streamlines the development process. A framework provides scalability, manages complexity, and maintains the modular design of a system. Currently, there is no framework that covers all stages of the development of a harmful detection system. For the first time, we addressed this challenge, which is discussed in Section 6.

**(3) The lack of detailed guideline for annotation**
In order to develop a harmful language detection model, unlabeled data are collected from social media. The data must be labeled in order to be used as input into machine/deep learning models. Annotation refers to the process of labeling data manually by some individuals. One of the major issues is that there is no standardized guideline that covers different situations. There should be a guideline for these individuals in order to ensure that the data is properly annotated. The advantages of the guideline include solving disagreements between annotators, coverage of specific conditions, and saving time and resources during annotation. In Section 6.3, we provide a detailed solution for this issue.

**(4) Considering the sentiment of the text**
Integrating sentiment analysis techniques into harmful language detection systems can improve the context sensitivity and effectiveness of identifying harm on online platforms. Sentiment analysis algorithms can detect these negative sentiments in text, helping to flag potentially offensive content. Section 6.3.1.2 addresses this issue.

**(5) Determining the source of harmfulness**
Identifying the origins of harmfulness is important for enhancing the interpretability and explainability of systems designed to detect harmful language. By pinpointing particular linguistic characteristics or contextual signals associated with harmfulness, these systems can offer more insightful explanations for their judgments. In Section 6.4, this subject will be discussed.

In addition, there are other challenges associated with harmful language detection as shown in Figure 1. Understanding harmful words for humans is intuitive; however, challenging to train a model to detect those. Considering that each comment has a high impact, spam review recognition and fake account tackle are important [97]. Awal and et.al. [98] study improved automatic spam users detection with more comments readability. The time period of extracting corpus is vital; for instance, using offensive language election period is a high rate, in addition in some cases, content changes over time [99]. Many previous works overcome this challenge by gathering data in different time periods [100], gathering a high amount of data [101], or using transfer learning [102]. Co-reference resolution is related to finding an individual or group that has been targeted offensive in the corpus in [103, 104]. In addition, derived fake correlation between word patterns and annotated labels makes the dataset biased, unreliable, imbalanced, and spars [105, 106]. Often expressed attitude does not contain offensive words although sarcasm, slang, emoji, or hashtag reverses the whole meaning of the text. As a result sarcasm or slang recognition challenges traditional NLP methods [107, 108].

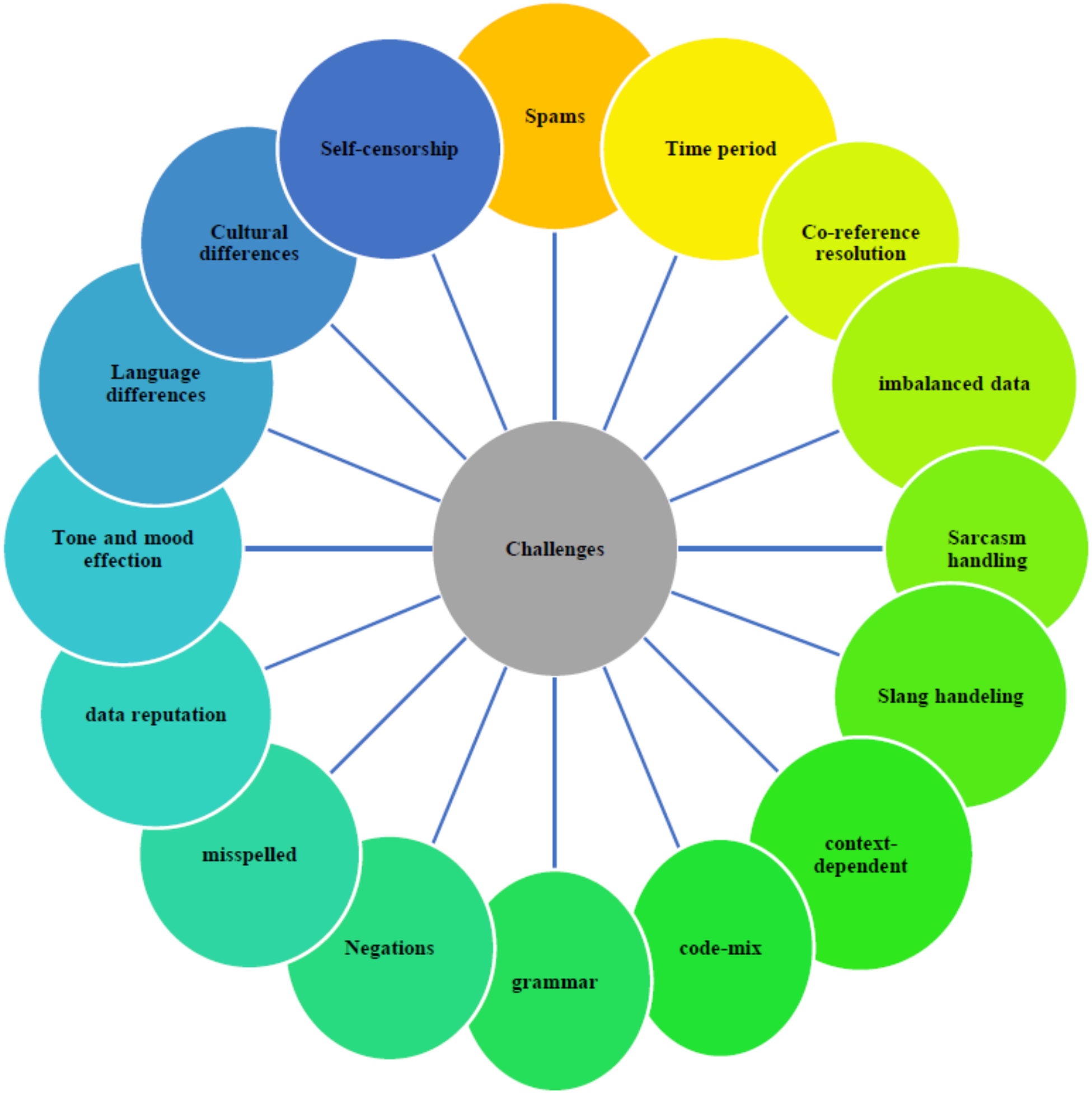

Figure 1: Harmful language challenges.

Most of the datasets in harmful language detection are provided on hate topics. Context dependency challenge refers to words meaning changing in different contexts [108, 109]. Domain and context of using a word can affect the whole analysis. The popularity of non-native social media made the intrusion of language to each other, especially English to other languages. Mixing two or more languages for interaction known as code-mix has affected languages' syntax, grammar, and morphology [110, 111]. Finally, these issue barriers to linguistic models and algorithms recognition. Using negative words or swearing in a corpus makes semantic analysis confused [112]. The existence of these words leads the model to misinformation and is hard to handle. the misspelling occurred often in fast or careless typing that make words not recognizable by algorithms [113, 114].

In order, other challenges in this field are, data reputation is originating from a lack of benchmarks or standard policy for datasets. In addition, a variety of techniques data crawling, and data gathering make it difficult to investigate the reputation of data [115], contrary to computer, human brain can understand para-linguistic features like mood and tone through the text interaction. Mood affecting offensive analysis text is just an emotion or attitude express and does not have any target [116]. Also text taken harmful subjectivity based on individual beliefs, and speculation [117]. Text not contain any offensive words instead transfer offense, and hate as tone, emotion and emoji to target [118, 119]. Language barrier makes harmful [120], culture diversity and social media culture is harmful [121], self-censorship especially effective in word-based recognition algorithms (such as A$$, etc.) [115].

## 5. Defining harmful language and other related concepts

Research studies have used a variety of expressions including offensive, abusive, hateful, cyber bullying, etc., which all convey negative content. In our study, all of these terms are grouped under a single category that we refer to as harmful language, as shown in Figure 2. One of the major challenges faced by researchers who attempt to classify harmful language in online platforms is the difficulty in defining related terms, which often overlap and can be interpreted by different individuals in a variety of ways. It leads to the existence of heterogeneous resources each reflecting a subjective perception.
Therefore, in the following, we have outlined some of the most common forms and their definition from difference resources to create a baseline for providing an accurate definition.

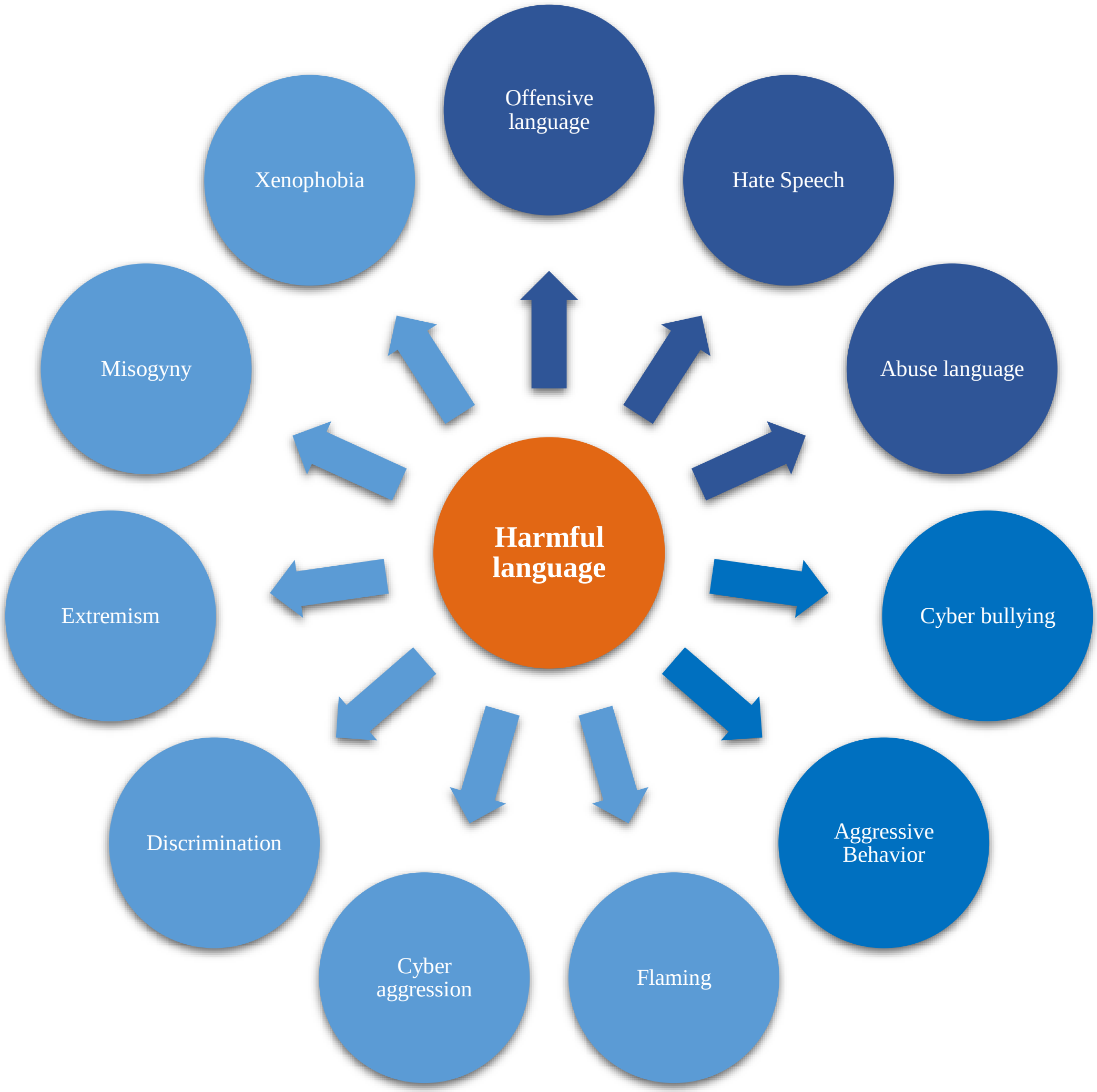


Figure 2: Various forms of harmful language.

### 5.1. Harmful content

Online harms represent a significant and escalating issue, with targeted groups and individuals enduring its consequences for an extended period. Harmful content manifests in various types, encompassing hate speech, offensive language, bullying, harassment, misinformation, spam, violence, graphic content, sexual abuse, self-harm, and numerous other types [123]. The UK government identified the following subcategories of harmful content: cyberbullying, dissemination of images and videos depicting child sexual abuse, cyber-flashing, propagation of terrorist group propaganda, and dissemination of disinformation and

misinformation [124]. It is worth noting that each online platform operates under its own set of content moderation policies and adheres to its own definition of harmful content [123].

### 5.2. Offensive language

The term offensive language refers to comments that are derogatory, hurtful, or obscene and can be related to other concepts that include abusive language, hate speech, cyberbullying, or toxic language [49]. The offense may take the form of inappropriate language (profanity) or a targeted offense, which may be disguised or direct [50]. Using fighting or hurtful language to insult a targeted individual or group with profanity, strongly impolite, rude or vulgar language [23]. A variety of offensive language include spreading hate, harassing others, writing aggressive or toxic comments, and posting or sharing offensive images and videos [17]. Chinivar et al. [125] considered any kind of abuse, hate speech, misogyny, xenophobia, troll, cyber aggression, cyber bullying offensive behavior.

As defined by OLID/OffensEval, offensive language consists of two elements: (i) language that is unacceptable and (ii) language that is targeted against a specific individual. There are three subcategories of offensive content considered by GermEval 2018 [49], including PROFANITY, INSULT, and ABUSE.

- **PROFANITY:** The text uses profane words, however, it does not intend to offend anyone. In some cases, profanity may be used in text that convey positive sentiments to emphasize the importance of a certain point.
- **INSULT:** There is no doubt that this text is intended to offend somebody and express disrespect and contempt. An insult is a negative evaluation of a person's qualities or lacks, or a label of unworthiness or under-valuement.
- **ABUSE:** The purpose of this type of degrading is to assign to a member of society an identity that is seen negatively by a majority of the population. The identity is viewed as embarrassing, unworthy, or morally objectionable.

### 5.3. Abuse language

Abuse occurs when someone causes harm or distress. An individual can be abused anywhere, including social media, messaging applications, email, gaming applications, live streaming websites, etc. [17]. A definition of abusive language would include hate speech, derogatory language, and also profanity [70]. According to Fortuna and Nunes [23], abuse may be defined as any form of strongly impolite, rude or hurtful language using profanity, showing degradement towards someone or something, or expressing intense emotions.

As described by Waseem et al. [106], abusive language can be classified into explicit and implicit forms. When abuse is explicit, it is unambiguous (such as racial slurs or homophobic remarks) and will often be identified by specific words. Those who use implicit abusive do not directly imply abuse, but instead use terms that are ambiguous, sarcastic, and do not contain profanity or hateful insults. This makes it more difficult to detect this type.

### 5.4. Hate Speech

The term "hate speech" refers to an expression of opinion expressed online intended to attack a particular person or group based on their gender, disability, ethnicity, religion, race, or sexual orientation, among other things [126] .

Among the two criteria Sanguinetti et al. [127] consider when determining hate speech are the action and the target. Actions include spreading hatred or inciting violence, or threatening the freedom, dignity, and safety of individuals. A target considered to be a protected group, or an individual targeted based on his/her membership in such a group, rather than on the characteristics of the individual. Based on our examinations, there is a consensus regarding the definition of hate speech, and other studies [23, 32, 128-130] have employed similar definitions.

### 5.5. Other terms

there are other terms that are related to offensive language including cyber bullying, cyber aggression, flaming, aggressive behavior, discrimination, extremism, misogyny, and xenophobia. Table 5 summarize these terms and provide their definition.

Table 5: Definition of different harmful language.

| **Common Types** | **Definition** |
|---|---|
| Offensive language | Expression of profanity, impoliteness, rudeness or vulgarity. |
| Abusive language | Intentionally harming or distressing someone. |
| Hate speech | Targeting a group or a person because of his/her membership in a group. |
| Cyber bullying | The act of abusing, embarrassing, intimidating, or aggressively dominating others by posting harmful, embarrassing, or threatening content is a form of cyberbullying. A victim who is unable to defend himself or herself is typically the subject of this kind of repeated and hostile behavior [72, 131, 132]. |
| Cyber aggression | Generally, online aggression occurs occasionally between peers without any intent to harm [133]. |
| Flaming | The use of profanity, anger, or hostility in comments that may disrupt community interaction and participation [134]. Flaming is directed at specific participants within a discussion [23]. |
| Aggressive Behavior | Intense, angry, and sometimes violent interactions with another individual or group of individuals via electronic means, intended to cause damage [135]. |
| Discrimination | Discrimination is a form of hate speech that assigns value to individuals on the basis of their race, ethnicity, or nationality [136]. During discrimination, distinctions are recognized and then used to justify unfair treatment [137]. |
| Extremism | Using political, religious, and/or social topics to segment society according to hateful ideologies [138, 139]. By using social media, they spread their ideologies, promoted their acts and recruited supporters [140]. |
| Misogyny | It refers to the hatred of women and sexist form of speech that targets women and keeps them at the bottom of society [141, 142]. |
| Xenophobia | unreasonable hatred towards foreigners [143]. |

### 5.6. Our definition of harmful language

It is not uncommon for offensive content to be referred to as abusive content in research articles since both use profane word. Studies have also shown that cyberbullying and hate speech content use profane language [4, 144]. Also offensive language, abusive language and hate speech have close relation mostly considered as equivalent or cause and effect [64, 145]. The distinctions between hate speech and offensive language was examined by [32], noting that these terms are frequently confused together.

A major difference between offensive and abusive language would be the inclusion of intentionality in abusive language, whereas the definition of offensive emphasizes lexical content and the receiver's emotional impact [59]. In this context, hate speech, which is directed at a specific individual or group, falls within the category of abusive language [146]. In [147] study, it is outline that aggressive behavior there is an intent to harm, where in [59] study, the intention is considered as abusive behavior.

The [125] study categorized all types as a subset of “offensive language”, while [122] regarded them as “abusive language”, and [147] study use “conflictual languages” to refer harmful, aggressive, abusive, and offensive languages, and other types. The [4] study suggested “detrimental/harmful content” includes hate speech, fake news, rumors, cyberbullying, toxic content and child abuse material. The [148] utilize

the term "harmful language" as a versatile term that may be substituted with expressions such as toxic, hate speech, and offensive language, among others.

It is evident that there is ambiguity and confusion in the definitions provided by studies, requiring serious attention. As stated by [10], a specific and well-structured terminology is essential for enhancing the effectiveness and practicality of automated solutions. It is important to differentiate between these categories to gain a clearer understanding of the connections among various phenomena and to develop improved language resources for each of them [59]. The successful development of machine learning models relies heavily on the presence of consistently labeled training data. However, a significant obstacle lies in the diversity of terms and definitions, leading to the interchangeable usage of related terms [122].

As part of contribution of this paper, drawing from our examination of various studies and their definitions, whether presented explicitly or vaguely, we have provided the following definitions for the most commonly employed terms:

- **Offensive language:** Using profanity and vulgar language to inflict **emotional** harm on the recipient.
- **Abusive language: Intentionally** using profanity and vulgar language to insult the recipient.
- **Hate speech:** Spreading or inciting hatred, discrimination, or violence towards individuals or groups based on attributes such as race, ethnicity, religion, gender, sexual orientation, disability, or other characteristics.

In addition, we present a definition for harmful language:

(1) All forms of offensive language, abuse language, hate speech, cyberbullying, cyber aggression, flaming, aggressive behavior, discrimination, extremism, misogyny, xenophobia, and other types which express negative content are categorized as harmful language.

(2) Covering all following words, phrases, tones, and feelings implicitly or direct: hurtful, hateful, malicious, rudeness, offensive, bad, cruel, harsh, aggression, insulting, improper, disparaging, violent, profane, derogatory, bullying (cyberbullying), profanity, discrimination, toxicity, flaming, obscene, swearing, vulgar, humiliating, mocking, dismissal, demeaning, belittling, slamming, personal attacks, nastiness, harassment, condescension, hostility, racism, slurring, radicalization.

(3) It is any content that is not appropriate for public and cannot be seen in front of family members and children.

We used the term “language”, because it implies nature of the sentence [147, 149]. As a result, we employ the term "Harmful Language" to prevent any ambiguity or confusion, referring explicitly to the definition outlined in this study.

## 6. Proposed cross-language framework

In order to detect harmful language, a framework can provide a consistent and standardized approach. By defining clear guidelines, rules, and procedures, harmful language can be identified and dealt with more efficiently. The result is a uniform and reliable detection process across different languages, contexts and applications. A proposed framework for harmful language detection is shown in Figure 3, which consists of seven main components. The first and second steps are to provide an input dataset and to preprocess it. Next is the annotation of the data, which requires a proper guideline in order to result in a more accurate classification. Vectorization and classification are performed once the data have been prepared. Lastly, the results are evaluated to determine the effectiveness of the approach used. To the best of our knowledge, this is the first comprehensive framework proposed for harmful language detection that covers all of the required components. This approach can be applied to any language, regardless of its resource availability (high or low resource).

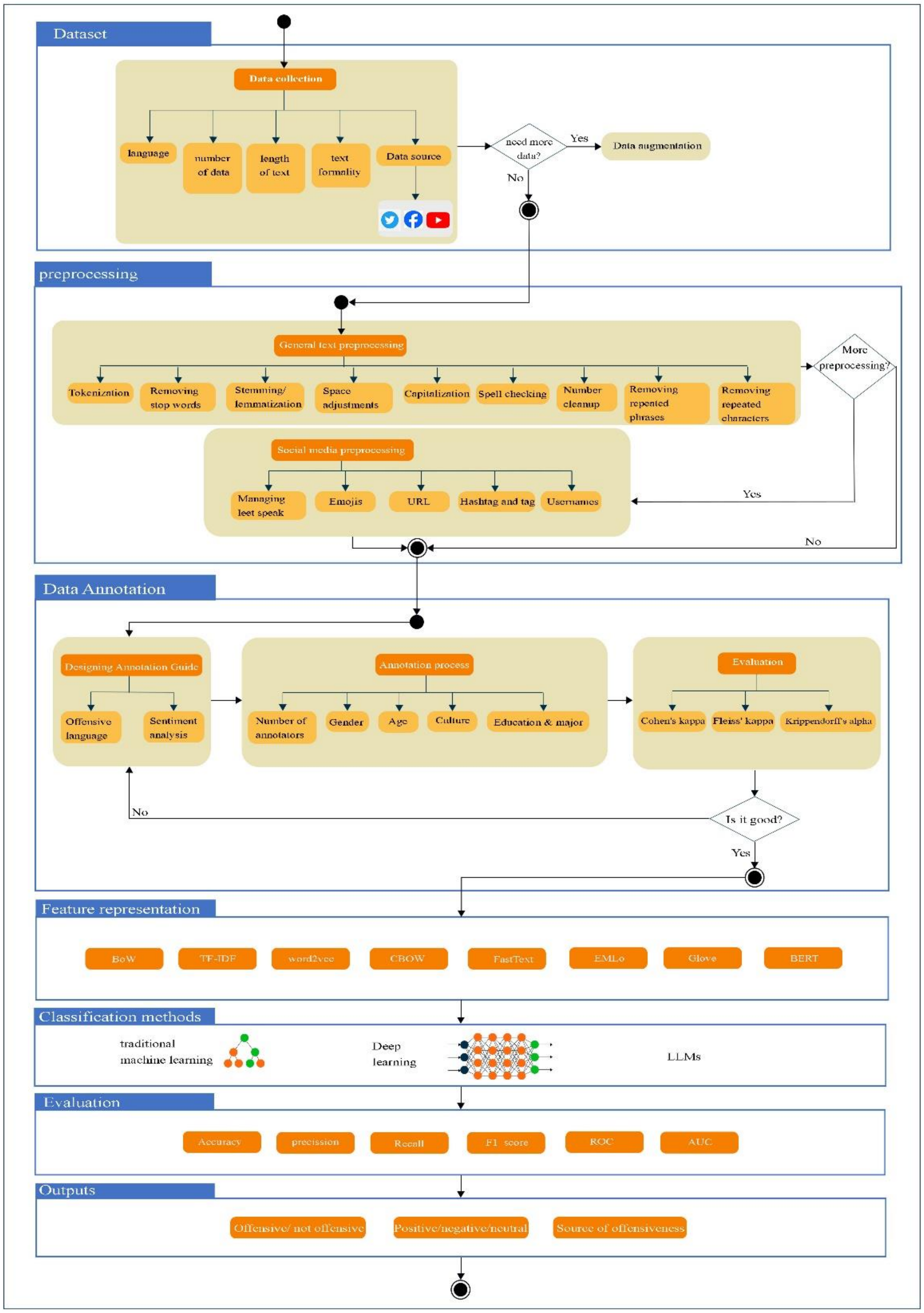


Figure 3: Proposed framework for harmful language detection.

## 6.1. Dataset

### 6.1.1. Data collection

Data quality and quantity are two important factors that should be considered during the data collection process [150, 151]. Data can be retrieved from a variety of online platforms. In the world of social media, Twitter was one of the most important platforms. It was provided a straightforward API for accessing data, making it a popular information resource for academic researchers, companies, and other organizations. It was therefore an ideal tool for studying social networks in a vast network with millions of members [152, 153]. Twitter has been found to be the most exploited source of data in previous studies, due to its relatively short text length and the fact that its data was publicly available [24]. It should be noted that Twitter recently changed its name to X[1] and changed its policies regarding data collection. Consequently, it is difficult to collect large amounts of data for the purpose of creating a suitable dataset. In addition, several other social media platforms such as Facebook, Reddit, Instagram, comments on newspaper articles, YouTube videos, and Wikipedia discussion forums may be considered as sources of information. Due to the fact that platforms typically report and remove harmful content to protect their users, it is difficult to create large datasets. Often, these datasets are small and do not include all forms of harmful language [30].

It is important to determine two factors in this step of the framework, which are the language of the text and the online platform from which data will be collected. Among the other parameters to be considered are the number of data, the length of the text (short text or long text), and whether it is formal or informal.

#### 6.1.2. Data augmentation

The term data augmentation refers to a set of algorithms used to generate synthetic data from an existing dataset. In most cases, this synthetic data consists of small changes in the data, which the model is supposed to be able to predict invariantly [154, 155]. It is beneficial to addressing some known problems, such as linguistic variation and overfitting risks [156]. The use of synthetic data is also capable of overcoming privacy issues associated with the use of social media data obtained without the consent of the user [157]. All deep learning models often require extensive datasets to attain satisfactory outcomes. Regrettably, acquiring sufficient data for real-world applications is not always feasible, and it is widely acknowledged that a lack of data readily leads to overfitting [158].

Data augmentation approach has been proposed as a method to improve the harmful language detection in several works [159]. It has been demonstrated by Casula and Tonelli [157] that generative data augmentation is effective in a variety of scenarios for harmful language detection. Furthermore, despite the abundance of comments found online, merely a fraction of them contain harmful content, leading to an unequal distribution between harmful and non-harmful data. This class imbalance can introduce bias in classifiers, favoring the class with a larger number of samples. To rectify this issue, various studies employ data augmentation techniques [160, 161].

### 6.2. Preprocessing

Preprocessing is an essential step employed to transform such free text into a structured form in order to effectively analyze it [4]. A major disadvantage of social media content is that it is noisy and written in an informal style [162], which makes analysis of the text difficult and reduce the performance of models. Following is a list of the important preprocessing steps [33, 163-165], specifically for harmful language detection, that should be followed for any type of text [166], which can be applied to any high/low resource languages [147].

- **Tokenization**: Tokenization involves breaking down the text into individual tokens, such as words or subwords. This step is crucial for subsequent analysis, as it allows the text to be represented as a sequence of meaningful units. Tokens can be identified by using spaces as delimiters or more

[1] We refer to it as X in the entire article.

sophisticated techniques like word segmentation. In some profane words, special characters are used like "f@@c", “sh*t”, which makes tokenization difficult [162].

- **Removing stop words:** Text analysis should eliminate stop words, such as articles and prepositions, that do not count as keywords and provide no useful information. It is important to identify the stop words that are unique to each language [76, 167].
- **Stemming and lemmatization**: Stemming and lemmatization are techniques used to reduce words to their base or root form. Stemming aims to remove common word endings, while lemmatization considers the context and converts words to their dictionary or lemma form. These techniques help in reducing word variations and standardizing the text [168].
- **Space adjustments:** It is required to replace sequences of two or more spaces with a single space. The tokenizer is also improved by inserting a space after the symbols like “!” or “?” [33].
- **Capitalization**: Based on the specific task requirements, the text can be converted to either lowercase or uppercase. Lowercasing the text can aid in standardization and reduce vocabulary complexity. In this way, there will be no difference between the letters of the same word conveying the same meaning but written in mixed case style. Using lower case on the entire text facilitates analysis and reduces data dimension [169, 170].
- **Spell correction:** It consists of identifying and correcting spelling errors in text data during text preprocessing [171].
- **Number cleanup**: By replacing all numbers with a single number, the dimensionality of vectorized texts will be reduced [33]. In other cases, numbers are removed [167].
- **Removing repeated phrases**: It is possible for the same word or phrase to appear repeatedly, which can give a greater significance to the words that are not as important. The first appearance of a word or phrase is retained, and repetitions are removed.
- **Removing repeated characters**: As a means of emphasizing a specific character, words may contain repeated character. As a result, the words are excluded from common vocabulary, and, consequently, pre-trained embeddings cannot assign them to vectors. It is therefore necessary to eliminate repetitions of characters [172].

In addition, there are some other preprocessing steps that are specific depending of data source. For example, a X may contain text, images, videos, URLs, hashtags and user mentions [152]. For this reason, some preprocessing steps are required for social media including:

- **Managing leet speak:** Leet involves the replacement of certain letters with numbers whose shapes are similar. Computer systems have difficulty interpreting this type of tweet. It is necessary to convert the numbers into equivalent predefined letters. An example would be the replacement of the text "n33d" with "need" [33, 173].
- **Emoji**: It is not uncommon for emojis to be converted to textual labels [121, 167] or replaced with their regular expression (regex) form in some cases, since they contain useful information regarding the emotions and intent of the user [174]. Some studies simply removed emojies [33] or replace it with the word “emojis” [175].
- **URL**: The word "address" [33] or “url” [172] can be substituted for detected URLs or links that did not provide valuable or meaningful information can be removed [176].
- **Hashtag and tag**: Hashtags can be replaced by the word like ‘‘label’’ [33] or “hashtag” [175], or in some cases be removed [76].
- **Usernames:** Usernames can be replaced by the word like ‘‘user’’ [33] or be removed [76].

Depending on language and its characteristics as well as data source, the above preprocessing steps that are appropriate can be chosen and applied to collected data. Python libraries including NLTK, Spacy, Gensim, and TextBlob are commonly used for the preprocessing of texts.

### 6.3. Methodology of data Annotation

Natural language processing relies heavily on labeled data. Data annotation, however, is a complex process and there are a number of valid opinions regarding what the proper data labels should be. Annotators' subjectivity has been acknowledged by dataset creators, but rarely actively managed during the annotation process [177]. While machine learning can be used to identify harmful language, one of the main challenge is annotating a sufficient number of examples to train the models [30].

An annotation may simply be a yes/no value indicating whether the text contains any harmful language. On the other hand, it is also possible to use multiclass labeling in which the type of harm is determined (offensive, abusive, hateful, etc.). One of challenges in this kind of approach is the definition of each type of harmful language, which we addressed in Section 5. Some studies used multi-level annotation [178] which is complex. There are several ways in which the labeling process may be carried out: by annotators with expertise, by non-experts and volunteers, through crowdsourcing platforms, or using automatic classifiers [24]. Annotation is time consuming and some studies (like SOLID [51]) used existing dataset (like OLID [50]) as seed to avoid it.

An annotation process can be made more efficient by having a guideline to aid in the process. This can assist annotators in adhering to consistent instructions, eliminating ambiguities, expediting the process, and enhancing accuracy. To the best of our knowledge, there is currently no general applicable annotation guideline available. This paper contributes significantly by designing a comprehensive guideline for annotating harmful language which can be adopted for any language. In the following, a detailed description of the elements essential for conducting a satisfactory data annotation is provided.

#### 6.3.1. Designing Annotation Guideline

A careful review of various studies was undertaken in order to establish a baseline for annotation based on efforts that have already been made in different languages. First, we need to establish some primitives at the beginning, which are defined in the following.

- **Data source:** Various data sources, such as text, images, and speech, can contain harmful language. In this study, we are interested only in examining text data. Moreover, the text data can be gathered from different online platforms.
- **Claim:** The text is either harmful or non-harmful.
- **Positive/ negative/ neutral tweet:** A feeling regard a text is intuitive. It is therefore beneficial to use emojis and hashtags in order to facilitate a better understanding of the tweet.

Next, some rules must be defined to follow during annotation, which makes the process more accurate. Therefore, false positive and false negative types (as discussed in other studies [179]) will be covered, and the error rate will be reduced. In this study, we considered both harmful language detection and sentiment analysis tasks. Thus, for each of them, an annotation guideline has been developed separately.

##### 6.3.1.1. Annotation guideline for harmful language detection

During the annotation process, we want to answer the most important and most challenging question: is a text offensive? Therefore, we considered two class namely harmful and not- harmful as shown in Figure 4. In order to categorize the text, we established the following criteria.

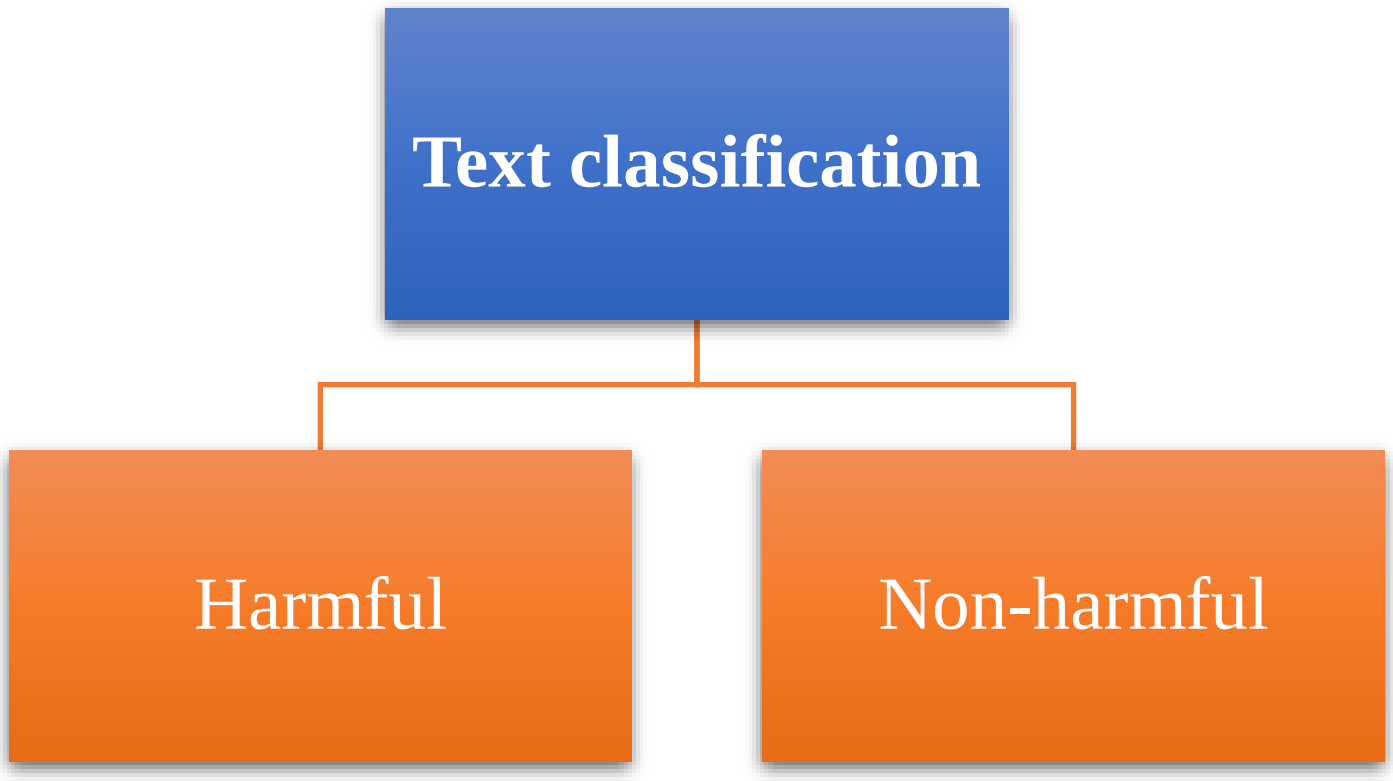


Figure 4: Text classification for harmful language.

**Class 1: Harmful**

The text is harmful if the following rules exist:

**Rule 1) The audience:** The text has a direct or indirect audience. In the direct case, the pronoun is clearly present in the text. In indirect case, the referent of the verb refers to a specific person, people, category, group or product.

**Rule 2) The context:** The text is a publishing, an invitation, or conveying a concept, feeling, opinion, viewpoint, evaluation, emotion or emotional interaction, all of which are negative, through the text.

**Rule 3)** The words in Table 6 or their synonyms be included directly or indirectly through word phrases, irony, sarcasm, tone, emotion, emoji, and hashtags.

**Conclude:** A text is harmful if a tweet meets all three rules. If it does not have one of the three items, the priority is 2 = < then 3 = < and then = < 1 in order.

Table 6: Harmful signs.

| | | | |
|---|---|---|---|
| hurtful | improper | obscene | nastiness |
| hateful | disparaging | swearing | harassment |
| malicious | violent | vulgar | condescension |
| rudeness | profane | humiliating | hostility |
| offensive | derogatory | mocking | racism |
| bad | bullying (cyberbullying) | dismissal | slurring |
| cruel | profanity | demeaning | radicalization |
| harsh | discrimination | belittling | troll |
| aggression | toxicity | slamming | extremism |
| insulting | flaming | personal attacks | misogyny |

It is important to pay careful attention to some special cases during annotation, which are listed in the following which are summarized in Table 7.

(1) The text has an audience. There is no bad word. It has the aspect of invitation, dissemination and transmission.
(2) The text has an audience. There is no bad word. The sentence is questioning (It has irony or colloquial expressions).
(3) The text has an audience or not. There is no bad word. The reference and meaning of the verb is negative and bad.
(4) The text has an audience. It can have a bad word or not. It has self-censorship. It has one of the three rules of being harmful.

The term self-censorship refers to the removal of certain words. By doing so, the writer removes a portion of a word or meaning from the text (like s&&k, bi&&h in English).

(5) The text does not have any of the harmfulness rules. But it has an audience. There is a bad word that makes the tweet negative.
(6) Harmful emoji is used.
(7) The text has an audience. It is ironic. The like symbol or positive emoji (eggplant, etc.) or laughter is used in a negative way.
(8) The text has an audience. There is no bad word. It has the mode of publication, invitation and negative transmission.
(9) The text has an audience. It can have a bad word or not. It has a publication, invitation, etc. aspect.
(10) The text has an audience. It can have a bad word or not. It has a publication, invitation, etc. aspects. It has a negative meaning. It has a negative tone.
(11) The text has an audience. There is no bad word. It has a negative meaning. Usually contain “?”, “!” or both.

Table 7: Harmful conditions.

| **Case** | **Audience** | **Bad word** | **Sentence type/ content/ mood** |
|---|---|---|---|
| 1 | ✓ | × | invitation, dissemination, transmission |
| 2 | ✓ | × | questioning |
| 3 | ✓/× | × | meaning of the verb is negative |
| 4 | ✓ | ✓/× | |
| 5 | ✓ | ✓ | |
| 6 | × | × | |
| 7 | ✓ | × | |
| 8 | ✓ | × | publication, invitation and negative transmission |
| 9 | ✓ | ✓/× | publication, invitation, etc. |
| 10 | ✓ | ✓/× | publication, invitation, etc. |
| 11 | ✓ | × | |

**Class 2: Non-harmful**

Text in the second category are non-harmful. Although identifying them may seem straightforward, it is not easy and there are some special cases that require extra attention. The following is a list of these cases, and Table 8 provides a summary of them.

(1) The text has an audience or not. It has a bad word. It does not have the aspect of invitation, publication or transmission. The mood of the sentence is news.
(2) The text has an audience. It has a bad word. A tweet is simply an expression of a state/event, or a quote or a memory.
(3) The text can have a bad word or not. It has negative emoji and hashtags. It has a news mode. It's just an emotion state announcement. It has publishing, transmission and invitation modes.
(4) The text has all three rules of harmfulness. But the audience is the writer himself/herself and the text becomes an expression of personal feeling.
(5) The text does not have any of the sarcastic conditions (3 rules). It is simply expressing a subject.

Table 8: Non-harmful conditions.

| **Case** | **Audience** | **Bad word** | **Sentence type/ content/ mood** |
|---|---|---|---|
| 1 | ✓/× | ✓ | news |
| 2 | ✓ | ✓ | state/event/quote/memory |
| 3 | | ✓/× | announcement |
| 4 | | | expression of personal feeling |
| 5 | | | expressing a subject |

### 6.3.1.2. Annotation guideline for sentiment analysis

Over the past few decades, techniques for tasks related to sentiment analysis have gained increasing importance within NLP [24]. A variety of purposes may be pursued including obtaining the views of users on a given product or surveying political opinions [180, 181]. Generally harmful language detection categorized by five level of text sentiment analysis granularity [120, 182].

- **Document level [183, 184]:** The task is to extract the main harmful attitude expressed in the whole document. Although cannot distinguish multiple harmful opinions.
- **Sentence level [185, 186]:** The task is finding harmful or not and subjectivity in each sentence expressed. Neutral opinions are not counted.
- **Aspect level [187]:** The task is finding harmful or not, subjectivity, and neutral opinion expressed in both document and sentence. Detecting target harmful category.
- **Comparative level [26, 188]:** Finding the indirect comparative expressed harmful relation and their entities in each sentence. This level of harmful language detection is suitable for small datasets.
- **Lexicon level [189-191]:** In this level, harmful words recognizing from an uncertain length text set. Lexicon level divided to feature a set of pre-defined words (dictionary), and Corpus a large set of text.

Various issues are associated with sentiment analysis and NLP, such as individuals informal writing style, sarcasm, irony, and language-specific challenges. There are many words in different languages whose meaning and orientation change depending on the context and domain in which they are employed [192]. For sentiment analysis, three annotators classified each sentence into one of three classes (Figure 5).

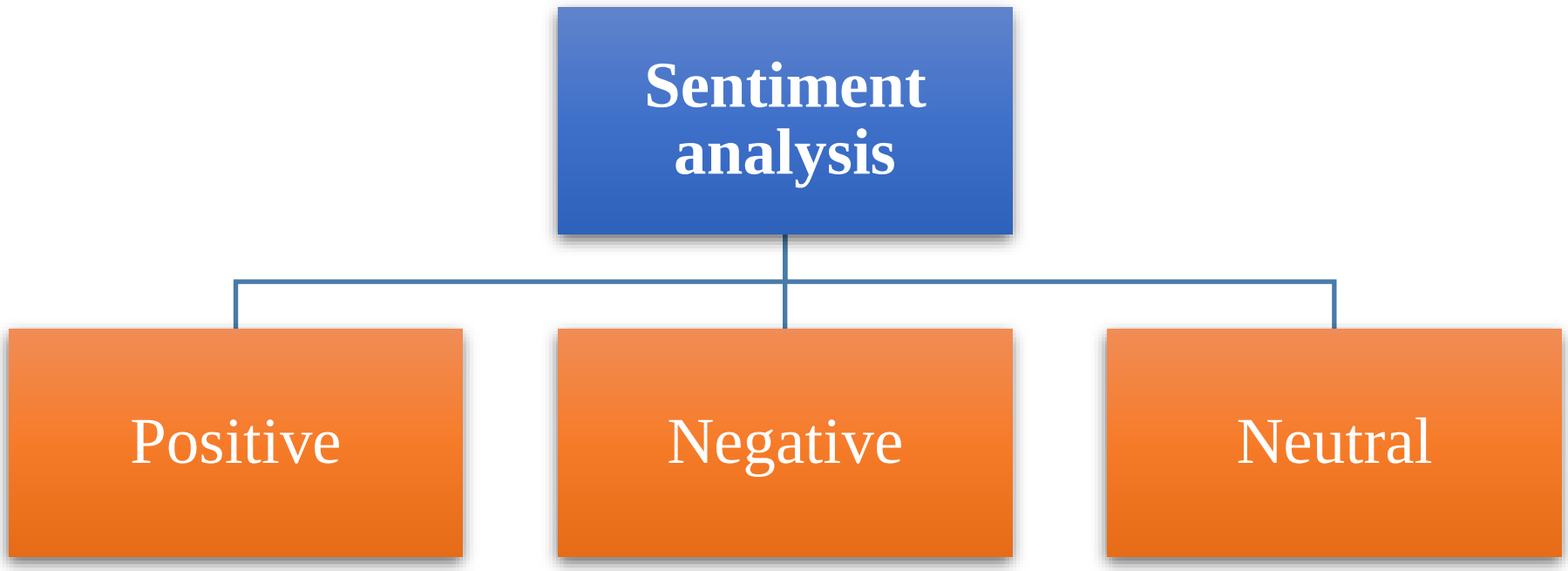


Figure 5: Sentiment analysis categories.

**Class 1: Positive**

- Whether explicitly or implicitly, the text suggests that the speaker is in a positive mood.
- A positive text means praise, praise, expressing positive feelings, and expressing satisfaction about a topic.

**Class 2: Negative**

- Whether explicitly or implicitly, the text suggests that the speaker is in a negative mood.
- It shows the audience's negative opinion about a subject.
- It means criticism, disgust, expressing negative feelings and expressing dissatisfaction.

**Class 3: Neutral**

- The text does not contain any indication of the speaker's emotional state, either explicitly or implicitly.
- Audience meaning is unclear and ambiguous

### 6.3.2. Annotation process

Annotation process involves finding individuals to label the dataset based on provided annotation guideline. Given the diverse backgrounds of individuals and their varying perspectives on harmful language, it may be necessary to consider certain criteria when selecting them. As harmfulness is subjective, taking these factors into account is essential. Unfortunately, most researches have not paid attention to or do not explain this issue clearly enough. To ensure the selection of a suitable diversity of annotators, it is crucial to take into account the following criteria:

- **Number of annotators:** During the initial stages of the process, this parameter should be determined. The number of annotators can range from one to many. Usually, an odd number is selected in order to resolve disagreements. In cases where there are a large number of annotators, tools such as Google forms can be helpful in labeling data.
- **Language:** It is important to determine the language of the annotator. The majority of the time, native speakers are selected.

- **Age:** Another parameter reported in studies is the age of the annotators. Due to the sensitive nature of harmful language, children and adolescents cannot participate in the labeling process. Furthermore, age diversity may lead to more disagreements between annotators, as different generations speak differently.
- **Gender:** The ratio of male to female annotators can be taken into account, as men and women may interpret different words in various ways. Typically, men tend to use harmful language more frequently, while women are generally more sensitive to it. Consequently, words that men may not consider harmful could be perceived as such by women.
- **Education and Major of study:** Well-educated annotators are more likely to comprehend annotation guidelines effectively. They can interpret instructions accurately and adhere to them meticulously, resulting in higher-quality annotations. There is no doubt that the major of an annotation can be an important parameter. Linguists can make better judgments about the potential harm conveyed by the text than other individuals.
- **Culture:** In some cases, the culture of the annotator can affect the results, as what may be considered harmful in one cultural context may not be perceived as such in another. It can be determined based on the nationality of the annotators.

#### 6.3.3. Annotation agreement and evaluation

An annotation often involves comparing annotations from multiple people on a text. In this way, annotation schemes and guidelines are validated and improved, ambiguities and difficulties in the source are identified, and valid interpretations are assessed [193]. In order to ensure data annotation accuracy, it is necessary to assess the degree of inter-annotator agreement, which measures how often annotators make the same annotation decision [194]. In order to verify consistency, we can perform the annotation process on the same source several times [193]. Inter-annotator agreement can be measured by a number of metrics, including Cohen's kappa, Fleiss' kappa, and Krippendorf's alpha.

**Cohen's kappa:** It indicates two annotators are in agreement beyond what might have been expected by chance [195]. It is calculated by using Equation 1, where *po* represents the predicted agreement between annotators, and *pe* represents the probability of random agreement [196]:

$$Cohen\ \text{kappa} = \frac{po - pe}{1 - pe} \quad (1)$$

**Fleiss' kappa:** An extension of Cohen's kappa that can be used when more than two annotations are involved [195].

**Krippendorff's alpha:** This measure can be used for incomplete data and also for scenarios where only partial agreement may exist between annotators [26]. It is calculated using Equation 2, where a weighted percent agreement is expressed as *pa*, while a weighted percent chance agreement is expressed as *pe* [197]:

$$\text{Krippendorff's alpha} = \frac{pa - pe}{1 - pe} \quad (2)$$

### 6.4. Why is it harmful?

In some scenarios, models based on machine learning are more accurate than human experts, but human moderators are more effective at detecting harmful language since they are able to explain their decisions. As machine learning models such as neural networks are black boxes, humans can identify harmful phrases or subtypes of harmfulness in the text. Explainable machine learning methods can be considered in order

to obtain more accurate answers and to explain what takes place in the model from input to output. Explaining machine-learned classifiers helps to build trust and increase their acceptance, which facilitates a fair and transparent moderation process [3]. Thus, in order to strengthen the originality of our proposed technique, we explore the root causes of harmful behavior which provide valuable inside regarding the outputs and results.

In more simple way, certain words like swear words can be seen as the source of harm in text. However, there are words that are usually not considered harmful, but depending on the context of a sentence, the writer may intend to be harmful by using such words. Therefore, as mentioned in [190], there are different type harmful words (Figure 6) which are explained in the following:

- **Animals:** In some cases, the name of an animal may be used as an insult (monkey, pig).
- **Body part:** Sometimes the name of a part of the body is used to offend someone (d&ck).
- **Celestial beings:** It may be a term that comes from a specific cultural, religious, or mythological context (seiten).
- **Conditions:** Mental disorder (crazy, idiot), sexual deviation (lesbian*)*, physical disability (blind, deaf).
- **Family member:** As a form of swearing, words that refer to relatives are used.
- **Occupation:** Religiously forbidden occupations, often used in offensive ways (servant, doorman).
- **Nationality:** A nationality that is used as an offensive word typically refers to some negative characteristic of the country. Sometime a name of city is also used.

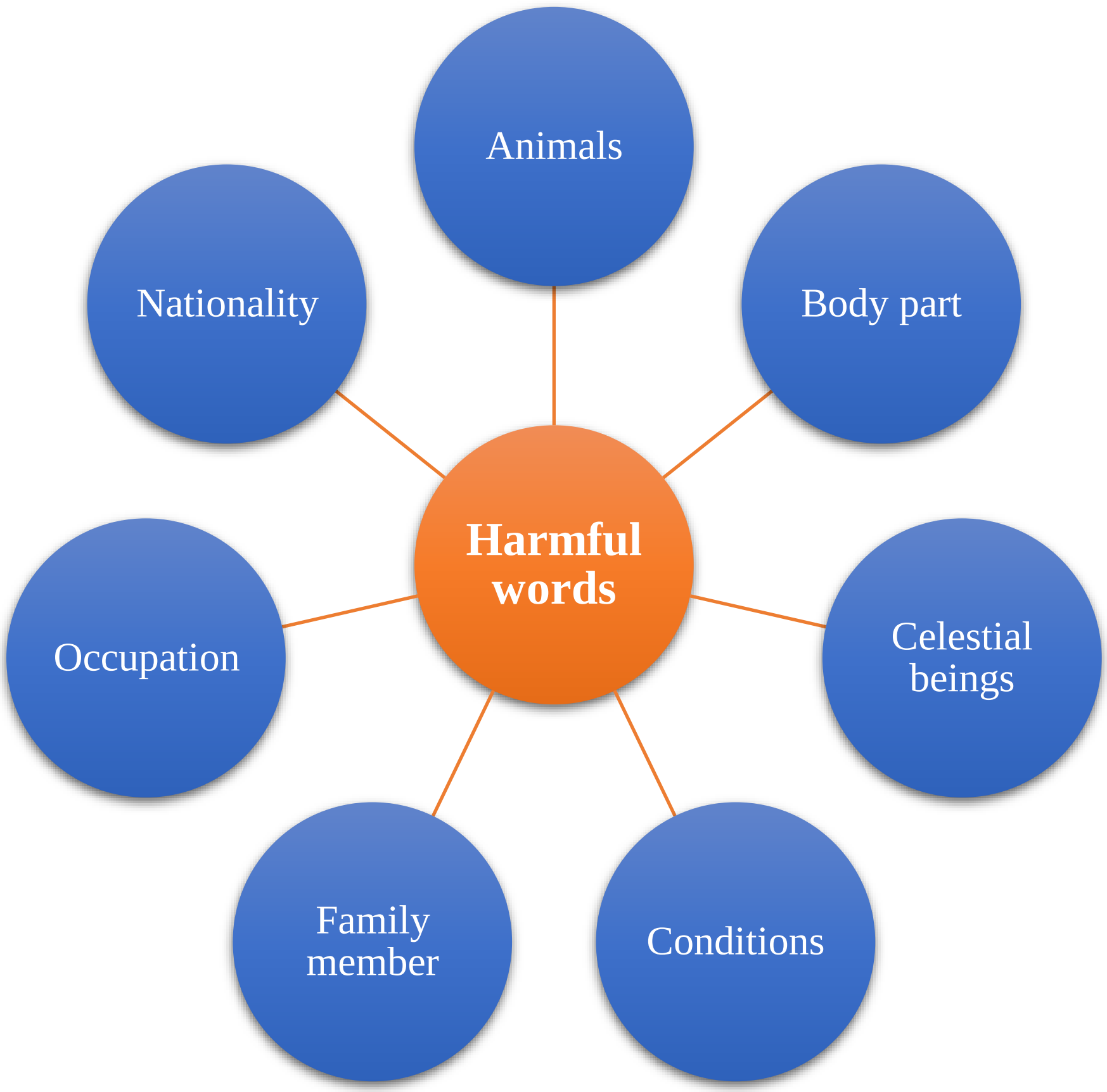


Figure 6: Category of harmful words.

### 6.5. Text representation

It is necessary to transform unstructured documents into structured data to become understandable by algorithms [198] and in order to conduct any type of text analysis [199]. In NLP, one of the primary steps involves converting the input text into a numerical format, such as a vector or matrix [200]. In word embedding, the semantic and syntactic meanings of a corpus are embedded in real-valued vectors [201]. Word embedding techniques are helpful in capturing contextual information much more effectively [202]. A variety of embedding methods are described below which are applicable in harmful language detection and applied in related studies.

**Bag of words (BOW):** It focuses solely on whether or not known terms appear in a document, rather than their location within the document. Through this method, all information regarding the sequence or structure of words in a document is eliminated [203].

**Term Frequency-Inverse Document Frequency (TF-IDF):** A statistical method for measuring the importance of words in a set of documents is the Term Frequency-Inverse Document Frequency. The total frequency of a word in a document is represented by TF, and its inverse frequency is represented by IDF [203].

**Word2vec:** Word2vec identifies both syntactic and semantic patterns in language by analyzing how words co-occur in similar contexts with neighboring words [204]. To achieve this, word2vec utilizes neural networks to map the connection between a word and its surrounding words within a set of documents [205, 206]. The Word2Vec tool provides two distinct neural network architectures, namely Continuous Bag of Words (CBOW) and Skip-gram. These architectures are shallow neural networks comprising a single hidden layer, designed to learn the vector representation of words [207].

**CBOW:** The CBOW model utilizes several words to represent a particular target word. This entails duplicating the connections from the input to the hidden layer $\beta$ times, where $\beta$ represents the number of context words. Consequently, the bag-of-words model is primarily employed to depict an unstructured group of words as a vector. Initially, a vocabulary is established, encompassing all the distinct words in the corpus. The shallow neural network's output focuses on predicting the word based on its context [208, 209]. The primary objective of the CBOW model is to train a word vector capable of predicting the central word within a given context [210].

**FastText:** The fastText approach is founded on the skip-gram model, in which every word is depicted as a collection of character n-grams [211]. Each character n-gram is linked to a vector representation, with words being represented as the summation of these representations [212]. The acquisition of word representation is achieved by considering an extensive context of preceding and succeeding words. fastText has the ability to generate an embedding for words that are misspelled, infrequently used, or absent in the training corpus, due to its utilization of character n-gram word tokenization [213].

**Global Vectors (GloVe):** GloVe closely resembles the Word2Vec approach, representing each word with a high-dimensional vector and training it based on the surrounding words in a large corpus. The widely utilized pre-trained word embedding is derived from a 400,000-word vocabulary and trained on Wikipedia 2014 and Gigaword 5, consists of 50 dimensions for word representation [214].

**EMLo:** ELMo is a context-based embedding that acquires contextualized word representations by relying on a neural language model that incorporates a character-based encoding layer and two BiLSTM. The character-based layer encodes a sequence of characters belonging to a word and produces the word's representation, which is then utilized by the subsequent two BiLSTM layers. These layers effectively employ hidden states to generate the ultimate embedding of the word [215, 216].

**BERT:** The BERT (Bidirectional Encoder Representation from Transformers) model utilizes the encoder component of the transformer to generate word representations. BERT is employed for constructing language representation models with diverse applications. BERT possesses an initial layer of "knowledge" acquired through pre-training. Building upon this foundational knowledge, BERT can undergo further training to adapt to specified requirements. BERT's transformer analyzes each word in a given sentence in relation to all other words, facilitating a comprehensive understanding of the word's contextual significance.

This stands in contrast to alternative models that merely comprehend a word's meaning in a singular dimension [217, 218]. This approach has been employed for harmful language detection task in numerous studies across various languages, such as German [219], Arabic [220], and Pashto [221], in recent years.

### 6.6. Classification

Text classification is a traditional problem in NLP aimed at assigning labels to units of text such as phrases, paragraphs, and documents [222]. The need for automatic text classification is becoming increasingly important with the increasing amount of text data being generated [223]. As reviewed in Section 2, there have been a number of recent efforts to detect different types of harmful language using machine learning, deep learning and recently large language models (LLMs). Here, machine learning models are explained as baseline. A baseline models, which usually lack complexity, serves as a reference for comparison purposes [224]. Then, an explanation about deep learning and LLMs is provided.

#### 6.6.1. Traditional machine learning methods

Traditional machine learning techniques have been extensively employed in the task of identifying different type of harmful language. The majority of approaches are based on supervised text classification tasks, typically necessitating feature engineering. Multiple researchers have integrated individual attributes like age, gender, and user activity history into their harmful language detection system [225, 226]. In these studies, following machine learning methods are used to classify harmful texts.

**Gaussian Naive Bayes:** Naive Bayes is relying on a probabilistic approach to make predictions. It can predict distinct features that are unrelated to others, simplifying the classification process. However, Naive Bayes only takes into account the probability of independent features, potentially leading to inaccurate predictions when the feature set in the training dataset is interdependent [227]. While Naive Bayes learners are efficient, they struggle with the limiting assumption of conditional independence between attributes [228]. Gaussian Naive Bayes relies on a probabilistic method that assumes each class have Gaussian normal distributions [229].

**K-Nearest Neighbor (KNN):** The purpose of employing the KNN concept is to forecast the closest similar datasets [230]. This method is commonly employed to classify a specific query not only based on its text, but also on the surrounding text region by using similarity-based learning approach [231].

**Logistic Regression:** Logistic regression analyzes the relationship between a dependent variable and independent variable(s) [232] and interprets maximum likelihood as the probability of an event occurring [230]. There are a number of basic assumptions in logistic regression, including the independence of errors, lack of multicollinearity, and the lack of significant outliers [233].

**Gradient Boosting:** Boosting algorithms merge weak learners, which are learners slightly more effective than random, into a strong learner through an iterative process [234, 235]. Gradient boosting is an algorithm similar to boosting used for regression [236].

**Decision Tree:** Decision trees are composed of internal and leaf nodes, in which internal nodes define the routing function and leaf nodes predict the class label. By recursively splitting leaf nodes, decision trees can be trained easily [237]. In addition to being extremely useful in a vast range of applications, decision trees are also known for their robustness and interpretability [217, 238].

**Random Forest:** The random forest is an ensemble learning approach for both classification and regression [239]. Multiple learning algorithms are used in ensemble methods to improve prediction performance. The

random forest method works by constructing multiple decision trees during the training phase [240] and use bagging [241] to determine classification output [242].

**SVM:** The process of training an SVM classifier entails identifying a hyperplane, serving as its decision boundary, to effectively separate instances with the greatest margin [243].

The disadvantage of traditional methods lies in the manual analysis of complex features, which can lead to the loss of critical information in detecting harmful language. Features play a pivotal role as they offer valuable insights into the harmfulness of text. To tackle this challenge, researchers are turning to deep learning methods, which offer automated feature extraction, thereby enhancing the efficiency and effectiveness of harmful language detection [244, 245].

#### 6.6.2. Deep learning methods

**CNNs:** CNNs have been successfully applied in harmful language detection tasks due to their ability to automatically learn discriminative features from text data [37, 246, 247]. These networks consist of several layers including convolutional layer to extract features, pooling layer to retrains most important features, and output layer to produce the output [248].

**Recurrent Neural Network (RNN) and its variants:** RNNs acquire knowledge from training input but are distinguished by their "memory," enabling them to influence current input and output by utilizing information from previous inputs. Nevertheless, conventional recurrent networks encounter the problem of vanishing gradients, which complicates learning with long data sequences. To address this issue, various variants such as LSTM, BiLSTM and GRU have been introduced, mitigating these challenges and demonstrating strong performance across numerous real-world applications including harmful language detection [248-250].

**Transformer-based model:** The Transformer has quickly emerged as the leading architecture for NLP, outperforming other models like CNNs and RNNs in different tasks. Its scalability with training data and model size, support for efficient parallel training, and ability to capture long-range sequence features contribute to its success [251, 252]. Over the past few years, transformer-based models like BERT [36], DistilBERT [253], RoBERTa [254, 255], XLM-RoBERTa [256], and others have gained recognition for their capacity to identify and categorize harmful texts through contextual learning.

Several studies have employed hybrid approaches like CNN-LSTM [207] and CNN-BiLSTM [257], leveraging the strengths of both models for categorizing harmful content. Some studies compared CNN, RNN, LSTM, and BERT models performance for offensive and hate speech detection [258, 259].

#### 6.6.3. LLMs methods

LLMs represent a distinctive category of pretrained language models (PLMs) achieved through the scaling of model size, pretraining corpus, and computational resources [260]. LLMs have been designed with billions of parameters and trained on trillions of tokens, offering the potential to tackle the challenges encountered by traditional text classification methods [261]. They represent a new opportunity for addressing the problem of detecting harmful language more effectively using Llama 2 [261] and GPT-3.5 [262] models.

### 6.7. Evaluation

Evaluation metrics play a crucial role in harmful language detection as they are assessing the performance of classification models. In this regard, confusion matrix (Table 9) functions as a structured table that provides a visual depiction of how well a model is performing by comparing predicted labels with actual labels [263, 264]. It breaks down the model's predictions into four categories:

- **True Positives (TP)**: Samples that the model correctly predicts as harmful.
- **True Negatives (TN)**: Samples that the model correctly predicts as non-harmful.
- **False Positives (FP)**: Samples that the model incorrectly predicts harmful when the actual label is non-harmful.
- **False Negatives (FN)**: Samples that the model incorrectly predicts non-harmful when the actual label is harmful.

Table 9: Confusion matrix.

| | **Predicted non-harmful** | **Predicted harmful** |
|---|---|---|
| **Actual non-harmful** | TN | FP |
| **Actual harmful** | FN | TP |

The following are a variety of standard evaluation metrics utilized in this field, applicable for assessing different classifiers.

**Accuracy:** In the total amount of predictions, the proportion of accurate predictions (see Eq. 1).

$$Accuracy = \frac{\text{true positive} + \text{true negative}}{\text{true positive} + \text{false positive} + \text{true negative} + \text{false negative}} \quad (1)$$

**Precision:** It is the ratio between elements correctly classified as True Positive, and all the instances classified as true (see Eq. 2).

$$Precission = \frac{\text{true positive}}{\text{true positive} + \text{false positive}} \quad (2)$$

**Recall:** It is the ratio between the True Positive and all the true elements (see Eq. 3).

$$Recall = \frac{\text{true positive}}{\text{true positive} + \text{false negative}} \quad (3)$$

**F1-score:** This metric combines precision and recall using the harmonic mean (see Eq. 4).

$$F1_score = \frac{2 * Precission * Recall}{Precission + Recall} \quad (4)$$

**Receiver Operating Characteristic (ROC):** The ROC curve is a graphical plot that illustrates the trade-off between true positive rate and false positive rate across various thresholds [265].

**Area Under Curve (AUC):** The AUC reflects the model's capacity to differentiate between classes. A model with an AUC score close to 1 indicates strong ability to accurately predict the intended class, while a score near 0 suggests an inability to distinguish between classes effectively. A score of 0.5 signifies that the model cannot differentiate between classes at all [265].

Commonly employed metrics in harmful detection researches include accuracy, precision, recall, and F1-score, with some studies also calculate ROC and AUC [266-268].

## 7. Implementation

In order to verify the effectiveness of our framework, we implement it for Persian language and evaluate its performance. The Persian language is a low resource language with its own challenges, making it an appropriate example for demonstrating the efficiency of the framework. Some of these difficulties include:

- **Linguistic features:** From a script perspective, Persian shares similarities with Semitic languages such as Arabic. However, linguistically, Persian belongs to the Indo-European language family [269], making it distantly related to many European languages and those spoken in the northern regions of the Indian subcontinent. These characteristics make Persian an intriguing subject for language technology studies [270].
- **Lexical ambiguity**: The Persian language is characterized by words that can have various meanings and interpretations depending on the context [271]. This ambiguity poses a significant challenge in precisely identifying harmful language, particularly when solely depending on lexical analysis.
- **Adjective usage ambiguity:** In Persian, adjectives can often replace nouns without any alteration in their lexical form, potentially leading to structural or semantic ambiguities in noun phrases [272].
- **Limited annotated data**: The availability of annotated datasets for detecting offensive language in Persian is restricted in terms of both size and diversity, especially when compared to languages spoken by larger populations. This scarcity of labeled data can impede the development of robust classification models.

Due to complexity of Persian language, we can argue that as long as the performance is acceptable in Persian, it can be assumed that it will be acceptable in any language as well. Here is an explanation of the implementation process, which follows our proposed framework steps.

### 7.1. Dataset

The first step of framework is data collection. The Persian language and X data were taken into consideration in this study. The dataset for this study was derived from the Dataak website [273], which provided 1.5 million tweets. A random sample of 30,000 tweets was selected, of which 2,368 contained harmful words and 27,632 did not contain any harmful words. Despite the fact that this data is similar to that used by [76], we have undertaken a different and more precise annotation process which follows the guideline explained in Section 6.3.1.1.
In compliance with [49], we also applied certain formal restrictions when sampling tweets, including:

- The tweets must be written in Persian.
- A tweet must contain a minimum of three alphabetic tokens.
- In tweets, URLs were not allowed since tweets with URLs often only become harmful when their linked content is considered.
- No tweet could be retweeted.

The purpose of all of these restrictions is to speed up the annotation process as much as possible. Figure 7 shows samples of tweets from the dataset.

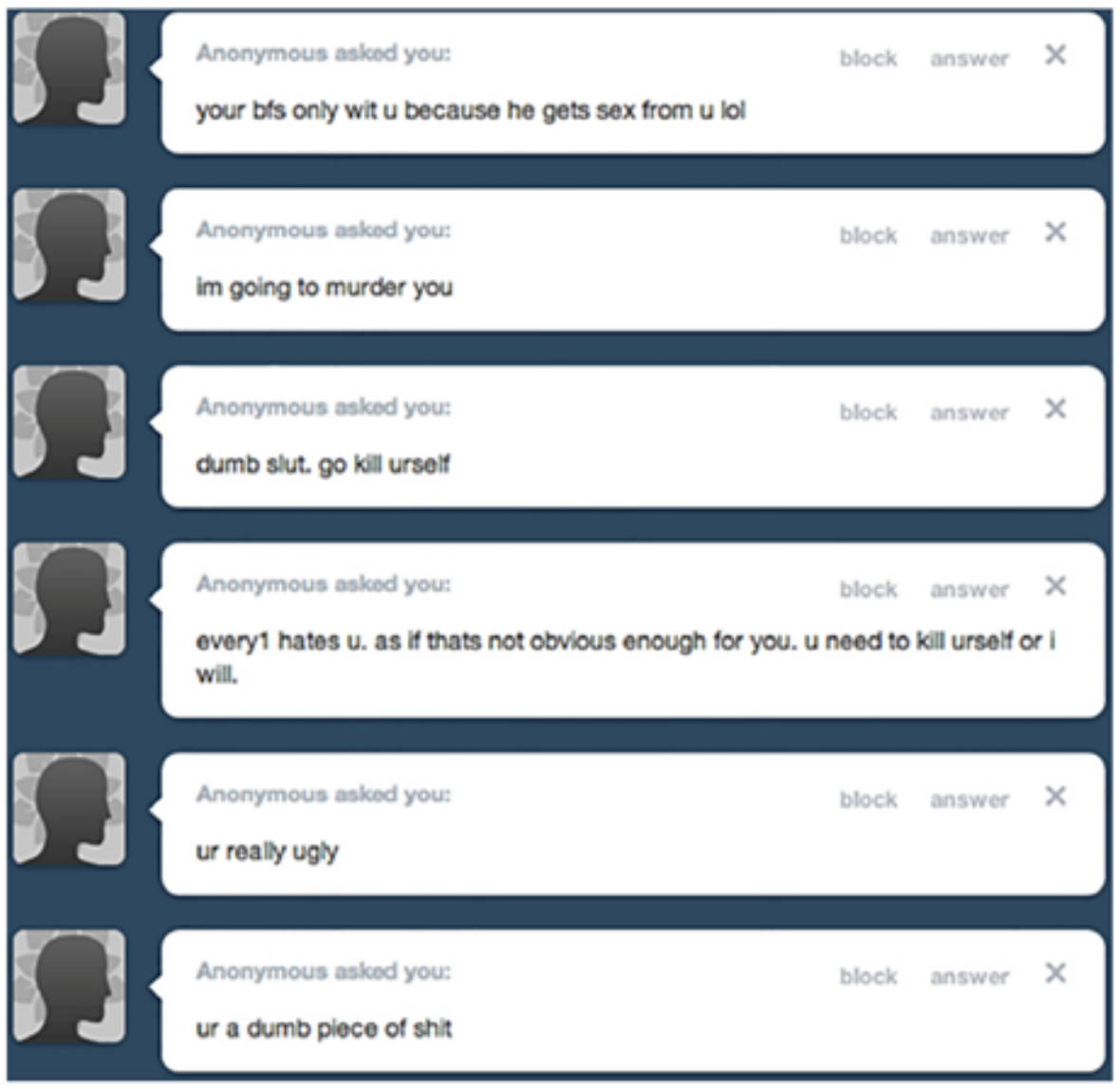

**Fig. 2.** Instance of an online abuse.

Figure 7: Sample of dataset.

### 7.2. Preprocessing

Different preprocessing steps was introduced in section 6.2. Some of these steps can be applied to Persian language, including tokenization, removing stop words [274], lemmatization, spell correction [275], number cleanup, removing repeated characters. Capitalization is one of the important steps in English preprocessing, but in Persian, there are no lowercase and uppercase letters, so this preprocessing is not applicable. Space adjustments is applied and additionally, Persian has a special kind of space known as half-space. It is recommended that all half-spaces be converted to spaces. Emoji where replaced with regex form. URL, hashtag and usernames were removed. Preprocessing in Persian is performed using two python libraries, Parsivar and Hazm.

### 7.3. Data annotation

Unfortunately, even though Persian is spoken by approximately 110 million people in Iran, Tajikistan, and Afghanistan [276], there are no detailed annotation guidelines for harmful language in Persian. Following our annotation guideline, three experts were responsible for annotating the data. Table 10 provides an overview of annotators characteristics. In appreciation of the time and effort they put into their work, the annotators were paid. We performed two annotation processes: one for the offensive language detection and one for the sentiment analysis.

Table 10: Annotators characteristics.

| **Criteria** | **Value** |
|---|---|
| Number of annotators | 3 |
| Language | Native Persian |
| Age | 28-30 |
| Gender | Male |
| Education and Major of study | • A master student in Persian linguistics.<br>• A doctoral student in Persian linguistics.<br>• A doctoral student in information technology and expert in NLP domain. |
| Culture (nationality) | Iranian |

### 7.3.1. Harmful language detection annotation

During harmful language detection, two categories were considered: harmful (with tag of 1) and non-harmful (with tag of 0). The sample sizes for harmful language detection are shown in Table 11. Almost 7.9% of collected tweets were harmful and 92.1% were non-harmful.

Table 11: Harmful language detection sample size.

| **Class** | **Size** |
|---|---|
| Harmful | 2,368 |
| Non-harmful | 27,632 |

To provide more accurate understanding of the conditions outlined in Section 6.3.1.1, Tables 12 and 13 shows examples of each scenario for both harmful and non-harmful text, respectively. There were 11 specific conditions for identifying harmful text and 5 specific conditions for determining non-harmful text, which required careful attention from annotators.

Table 12: Examples for harmful conditions.

| **Case** | **Example (Persian)** | **Translate (English)** |
|---|---|---|
| 1 | دانشگاه شما دانشگاه نیست | |
| 2 | الان مثلا باید بهت بگیم مهندس؟ | |
| 3 | رقیق تر بخور.<br><br>(مرجع مستقیم خوردن مشخص نیست – مفهوم و معنی بد دارد مرجع رقیق<br><br>تر به گه خوری بر می گردد). | |
| | کردن رفت.<br><br>(کردن فعلی است با معنی مثبت و منفی در این مثال کاملا به تکیه بر<br><br>اصطلاحات عامیانه و مفهوم ومعنی بد تشخیص صورت می گیرد). | |
| 4 | ملی(کا )زر نزن  ایموجی | |
| 5 | فاک می (: تهشه دیگه | |
| 6 | ایموجی | |
| 7 | بابا بذارید تو حال خودش باشه ایموجی | |
| 8 | زیباست .بازم پرواز تاخیر داشت. | |
| | (سازمان هواپیمایی مخاطب تمسخر و عدم رضایت است). | |
| 9 | از سری بعد خلاصه میگی کونی ایموجی<br><br>(قضاوت منفی(تمسخر وتنبیه )و قصد همراه کردن دیگران را دارد). | |
| 10 | کیر خودشه چیکار داری<br><br>(قضاوت منفی(تمسخر وتنبیه )و قصد همراه کردن دیگران و انتشار را دارد). | |
| 11 | وای یا خدا !این دیگه کیه؟<br><br>(قضاوت منفی(تمسخر وتنبیه )و قصد همراه کردن دیگران و انتشار را دارد). | |

Table 13: Examples for Non-harmful conditions.

| Case | Example (Persian) | Translate (English) |
|---|---|---|
| 1 | شت شت یه روز میریم | |
| | پشمام | |
| | آقا این میگه بن مژه!به والله ما یکی از دخترای گروهمون رفته بود انجام داد بقیه بهش میگفتن چرا و اینا!و من پشمام نیست از وقتی اینو دیدم ایموجی | |
| 2 | +"رفتی" تو میگم، چی نکردی دقت نه، -... چون رفتم، اره، +ولی رفتی تو بلند صداشو(<br>اره +بود بخودت حواست چون همین، -میرفتم بگا نمیرفتم اگه چون رفتم اره؛ )میکند حواسم به خودم بود، تو حواست به خودت نبود؟ | |
| 3 | آنفولانزا گرفتم ایموجی | |
| 4 | عجب ابلهی هستم. | |
| 5 | گام به گام | |

### 7.3.2. Sentiment analysis annotation

A tweet was assigned one of three labels in sentiment analysis: positive (+1), negative (-1), or neutral (0). Table 14 provides the sample sizes for sentiment analysis. From 30,000 collected tweets, 34.91% labeled as positive, 13.5% as negative, and 51.59% as neutral. Most of tweets have neutral or positive sentiment. Table 15 shows example of each class.

Table 14: Sentiment analysis sample size.

| Class | Size |
|---|---|
| Positive | 10,473 |
| Negative | 4,060 |
| Neutral | 15,467 |

Table 15: Sentiment classes examples.

| Class | Example (Persian) | Translate (English) |
|---|---|---|
| Positive | دمتون گرم، عالی بود | |
| Negative | استرس داشت، خوب بازی نکرد | |
| Neutral | چه لباسیم پوشیده | |

### 7.3.3. Annotation agreement and evaluation

We asked annotators to pay attention not only to the words used in a tweet, but also to its context. We emphasized to the participants that using a particular word does not necessarily mean that a tweet is harmful. The tweets were annotated separately by three individuals, and the final label was assigned according to the majority decision.

Using Fleiss' kappa, the inter-annotator agreement score in our annotation was 85% for the first time. To provide more accurate annotation, the process was reviewed and performed again. In second time, the score of 94% was achieved.

## 7.4. Why is it harmful?

To further enhance the novelty of our proposed method, we identify the reason for harmfulness. In this case, keywords are used to identify the root cause of harmfulness. There are some swear words in every sentence, and we identify them in order to clarify why the tweet was harmful. Therefore, we decided to search for additional keywords to make a more comprehensive list. In order to find other published lists of Persian harmful words, we searched the Internet. Using our extracted words from tweets (217 keywords) and lists provided by others like [277], the final list of harmful Persian words was created with total 479 keywords.

We present several samples of datasets in Table 16 and determine their harmfulness and sentiment labels. The source of harmful samples is also identified.

Table 16: Example of dataset, their labels and source.

| **Tweet (Persian)** | **Tweet (English translate )** | **Harmful** | **Sentiment** | **Source of harm (Persian )** | **Source of harm (English translate )** |
|---|---|---|---|---|---|
| نمیدونم چیه ولی آره | | 0 | 0 | - | |
| خوبه ببین😅 | | 0 | 1 | - | |
| وای بهار من سکته میکنم 😭😭😭😭😭😭😭😭 | | 0 | -1 | - | |
| تو ۴ تیکه استخون باشولی ۲ اصل مهم ( جیگر ، خایه )اینارو داشته باش، جثه ات دیگه مهم نی | | 1 | 0 | خایه/ | |
| دهنت سرویس بهترین توهین روز بود😂😂😂 | | 1 | 1 | دهنت سرویس | |
| بیا نصف میکنیم. نصف هزینه های خودش و دورو بریهاش هم با من 🤦‍♀️ اما دهنش بینده از طرف دخترا گه نخوره 😑 | | 1 | -1 | گه | |

## 7.5. Text representation and classification

The classifier we developed for harmful language detection and sentiment analysis was developed using different models with two main objectives: (1) demonstrating the effectiveness of our framework using a real low resource language dataset and (2) providing a baseline to be used in future research.

Hence, simple techniques were employed to establish a baseline in identifying harmful language, utilizing commonly applied machine learning models as detailed in section 6.6.1, coupled with TF-IDF for vectorization. It is worth mentioning that for SVM, a fast implementation of the SVM algorithm [242], called support vector classifier (SVC) is used.

Moreover, a deep learning approach is suggested, which is hybrid of CNN and LSTM architectures. In this proposed network, the initial layer involves embedding with randomized weights. A one-dimensional CNN is employed in the second layer for feature extraction. The third layer incorporates MaxPooling1D to identify important features. Following this, an LSTM layer is included to provide further insight about the context by considering both preceding and succeeding words. Lastly, a dense layer is added to generate output. The parameters of the CNN-LSTM model are outlined in Table 17.

Table 17: Training parameters of deep learning model.

| **Parameter** | **Values** |
|---|---|
| Epochs | 20 |
| Batch size | 8 |
| Learning rate | 0.0001 |
| Loss | Binary Cross-Entropy |
| Optimizer | Adam |

### 7.6. Evaluation and results

As shown in Table 18, deep learning based models outperformed traditional machine learning models in the detection of harmful language. According to the results, the deep + random weights model performed extremely well with an accuracy, recall, precision, and F1-score of 99.4%. A majority of machine learning models achieve similar outputs, but gradient boosting achieved better results with an accuracy and recall of 99.2%, and precision of 90.2%. As gradient boosting combines weak learners, it is capable of producing high predictive accuracy.

Table 18: Offensive language detection results.

| **Model** | **Accuracy** | **Recall** | **Precision** | **F1-score** |
|---|---|---|---|---|
| Gaussian Naïve Bayes | 0.772 | 0.772 | 0.876 | 0.815 |
| KNN | 0.919 | 0.919 | 0.868 | 0.884 |
| Logistic Regression | 0.921 | 0.921 | 0.848 | 0.883 |
| Decision Tree | 0.919 | 0.919 | 0.881 | 0.888 |
| Gradient Boosting | 0.922 | 0.922 | 0.902 | 0.886 |
| Random Forest | 0.920 | 0.920 | 0.879 | 0.886 |
| SVC | 0.921 | 0.921 | 0.878 | 0.884 |
| **Deep + random weights** | **0.994** | **0.994** | **0.994** | **0.994** |

A comparison of the performance of different sentiment analysis models is presented in Table 19. This multiclass classification was composed of three categories: positive, negative, and neutral. With an accuracy and recall of 66.2%, a precision of 67%, and an F1-score of 65.9%, the deep + random weights model achieves the best results.

Table 19: Sentiment analysis results.

| **Model** | **Accuracy** | **Recall** | **Precision** | **F1-score** |
|---|---|---|---|---|
| Gaussian Naïve Bayes | 0.515 | 0.515 | 0.508 | 0.457 |
| KNN | 0.528 | 0.528 | 0.488 | 0.436 |
| Logistic Regression | 0.548 | 0.548 | 0.540 | 0.444 |
| Decision Tree | 0.540 | 0.540 | 0.527 | 0.442 |
| Gradient Boosting | 0.547 | 0.547 | 0.567 | 0.445 |
| Random Forest | 0.538 | 0.538 | 0.521 | 0.444 |
| SVC | 0.544 | 0.544 | 0.546 | 0.446 |
| **Deep + random weights** | **0.662** | **0.662** | **0.670** | **0.659** |

## 8. Discussion

As a result of social media's potential risks, young generation mental health has been negatively affected, resulting in addiction, inattention, aggressive behavior, depression, and suicides [278]. Consequently, many attempts have been made to detect harmful language. For the purpose of comparing our proposed method with others, we considered some criteria including (1) the study worked on harmful language detection, (2) studies conducted on a variety of languages (high resource/ low resource), and (3) Provide guideline or follow existing one. In this way, comparisons are made with various articles with different features.

In view of the fact that OLID [50] is one of the foundations for annotations, we will compare our method with this article first. The main contribution of this paper is the development of hierarchical guidelines for offensive detection and the determination of its type and target. However, they do not provide any guidance regarding exceptions. Sentiment analysis is not performed and no framework is developed.

OMCD [81] dataset collected from YouTube was annotated according to [182] and no guidelines were provided. One interesting aspect of the dataset is that there are more offensive comments (4,304) than non-offensive comments (3,720). They did not remove emojis from the dataset and considered them to be part of the text in the same way that we did. The use of emojies can provide valuable information on harmful language detection.

DravidianCodeMix [26] considered both offensive language detection and sentiment analysis as we did. Their dataset is multilingual which followed OLID guideline for annotation.

For Persian language, Kebriaei et al. [80] focused on both offensive language and hate speech in their study. On the basis of the article, it appears that they provided a list of keywords. As mentioned by [129], keyword-based detection models may fail to (1) detect offensive in text without such keywords and (2) struggle to detect offensive in unseen data.
Ataei et al. [77] did not provide a clear definition of offensive language, often referring to it as abuse. In addition to following the OLID guidelines, more details were added. Compared to our dataset, there are fewer samples in their dataset.
A summary of the comparison with other studies can be found in Table 20.

Table 20: Comparing with previous articles.

| Article | Providing clear definition of harmful and its types | Proposing a framework | Providing annotation guideline | Detailed data annotation guidelines | Creating baseline | Sentiment analysis | Determining source of harmfulness | Best result (F1-score) |
|---|---|---|---|---|---|---|---|---|
| [50] | × | × | ✓ | × | ✓ | × | × | 80.00 |
| [81] | × | × | × | × | ✓ | × | × | 84.02 |
| [26] | × | × | × | × | ✓ | ✓ | × | Tamil: 74%<br>Malayalam: 94%<br>Kannada: 64% |
| [77] | × | × | × | ✓ | ✓ | × | × | 89.57 |
| [80] | × | × | × | × | × | × | × | 90.30 |
| Ours | ✓ | ✓ | ✓ | ✓ | ✓ | ✓ | ✓ | 99.40 |

## 9. Conclusion and future direction

It is widely acknowledged that harmful language is one of the dark sides of social media that has a detrimental effect on the user experience and quality of service. The issue has been addressed in a variety of ways in order to provide users with a more positive and inclusive digital space. In this study, we provided an overview of various proposed methods of detecting harmful language and reviews the state of the art in this field. Following that, we conducted a survey of existing datasets for different languages. Our review of these articles led us to encounter several important challenges, which we provide solutions to.
Among the main contributions of our paper is the development of a cross-language framework for harmful language detection. The framework consists of seven stages that should be determined in order to create a comprehensive system. Moreover, we provide a universal definition of harmful language so that it is clear what constitutes harmful and what does not. The following step was to develop a harmful language annotation guideline. The source of harm was also determined in order to gain an understanding of the reason for the harmfulness.
The Persian language dataset was prepared and annotated in accordance with our guidelines in order to assess the effectiveness of our framework. In order to create a baseline, different machine learning and deep learning methods are used. We demonstrate the effectiveness of our proposed method by harmful language detection with 99.4% accuracy, and sentiment analysis with 66.2% accuracy on a language with a low resource and a challenging dialect (Persian).

The harmful language detection still has some challenges that must be addressed in future research:

- **Lack of dataset for many languages:** Most of the collected datasets relate to the English language, and more datasets are needed, especially for low resources languages.
- **Automated annotation method:** Currently, annotations are performed manually in different languages, which is a time-consuming and error-prone process. It is necessary to develop an automated annotation model in order to eliminate bias and speed up the annotation process.
- **Identifying personal writing characteristics:** Since everyone has a unique writing style and the way in which they harm others, identifying a personal style can help create more accurate models.

- **Developing multimodal approaches:** Multimodal approaches are developed by considering different types of content, such as text, image, and speech. Because text and videos are frequently shared on social media, multimodal approaches can be useful in detecting harmful language.
- **Real-time offensive detection:** Different models in this topic are proposed based on collected data. Developing a model that can analyze social media posts in real-time and take appropriate action against harmful posts is essential.
- **Code-Mixing**: The difficulty of code-mixing in detecting harmful language occurs when individuals combine various languages or linguistic forms in their communication. This occurrence is prevalent in communities with multiple languages or on online platforms where users frequently transition between different languages. This code-mixing phenomenon introduces additional challenges for automated detection systems, as they need to handle multilingual input effectively.
- **Evolution of Language**: Language evolves over time, and new slang terms, expressions, or offensive phrases may emerge rapidly, making it challenging for detection models to stay up-to-date and generalize effectively.